\documentclass{article}
\usepackage[T1]{fontenc}
\usepackage{iclr2027_conference,times}

\usepackage{amsmath,amssymb,mathtools}
\usepackage{booktabs,multirow,tabularx,array}
\usepackage{graphicx}
\usepackage{float}
\usepackage{placeins}
\usepackage{microtype}
\usepackage{enumitem}
\usepackage{url}
\usepackage{hyperref}
\usepackage{xcolor}
\usepackage{tikz}
\usetikzlibrary{arrows.meta,positioning,calc,fit}

\newcommand{\cH}{\mathcal{H}}
\newcommand{\cA}{\mathcal{A}}
\newcommand{\cG}{\mathcal{G}}
\newcommand{\cD}{\mathcal{D}}
\newcommand{\KL}{D_{\mathrm{KL}}}
\newcommand{\ours}{\textsc{Joint Edit--Experiment}}
\newcommand{\indep}{Falsification + independent}
\newcommand{\E}{\mathbb{E}}

\newcommand{\argmin}{\mathop{\mathrm{arg\,min}}}

\definecolor{ink}{RGB}{45,45,45}
\definecolor{softblue}{RGB}{226,238,248}
\definecolor{softred}{RGB}{249,228,224}
\definecolor{softgray}{RGB}{242,242,242}
\definecolor{edgeblue}{RGB}{76,114,176}
\definecolor{edgered}{RGB}{196,93,79}

\title{Learning to Outgrow a Theory: Experimental Discovery Beyond the Initial Hypothesis Space}

\author{
\begin{minipage}{0.96\textwidth}
\centering
\normalfont
SiYuan Ma$^{1}$ \quad
Albert Gao$^{2}$ \quad
Chunzheng Zhu$^{3}$ \\[0.3em]
Xin Yan$^{4}$ \quad
Wenlong Zhang$^{5}$ \quad
Wenxin Zhang$^{6}$ \\[0.3em]
Luqi Gong$^{7}$ \quad
Tianlin Li$^{8}$ \quad
Qixin Zhang$^{1}$ \\[0.8em]
{\small
$^{1}$Nanyang Technological University \\
$^{2}$Carnegie Mellon University \\
$^{3}$Hunan University \\
$^{4}$Beijing Normal University \\
$^{5}$University of Science and Technology of China \\
$^{6}$University of the Chinese Academy of Sciences \\
$^{7}$Zhejiang Lab \\
$^{8}$Beihang University
}
\end{minipage}
}

\iclrfinalcopy

\begin{document}
\raggedbottom
\maketitle

\begin{abstract}
Scientific discovery systems typically optimize experiments within a fixed hypothesis space. 
This creates a failure mode when all available candidates omit the same missing mechanism: candidate disagreement can collapse even while the model class is systematically wrong.
We formulate \emph{experimental model-class revision}, in which a discovery policy jointly proposes a structural edit and a diagnostic experiment that tests whether that edit is necessary. 
The method couples a class-level distinguishability objective, in which one shared parameterization must explain all selected experiments, with anytime-valid sequential evidence that triggers structural revision only after the current class is rejected. 
On 400 held-out controlled dynamical environments, the joint policy reaches 89.5\% exact recovery with a budget of 32 real experiments, improving the strongest matched baseline by 10.0 percentage points while requiring fewer executed experiments and candidate fits. 
The learned revision–experiment pairing transfers across unseen mechanism combinations, held-out but expressible primitives, parameter extrapolation, and shifted experiment costs; when the true mechanism is outside the edit grammar, it detects library insufficiency in 88\% of cases with a 5.5\% false-support rate.
Revision gains also transfer to ODEBench and ODEBase model-library tasks, as well as DiscoverPhysics worlds. 
These results support a view of scientific discovery in which deciding \emph{what mechanisms a theory should make expressible} and \emph{where to collect evidence} are treated as a single sequential decision problem.
\end{abstract}

\section{Introduction}
\label{sec:intro}

Automated scientific discovery increasingly closes the loop between hypothesis generation, model fitting, and experiment selection \citep{1,2,3,4,8,44,46,47}. Most existing systems, however, assume that the current hypothesis space is sufficiently expressive, so informative experiments can be identified by comparing predictions among available candidates.

This assumption breaks under \emph{shared structural misspecification}, when all candidate models miss the same underlying mechanism. Although the candidates may differ in parameters or algebraic form, they can still make similar predictions because they share the same structural omission. As a result, low disagreement does not necessarily mean that one of the candidates is correct; it may instead mean that the entire hypothesis space is missing an important mechanism.

\begin{figure}[t]
\centering
\includegraphics[width=0.95\linewidth]{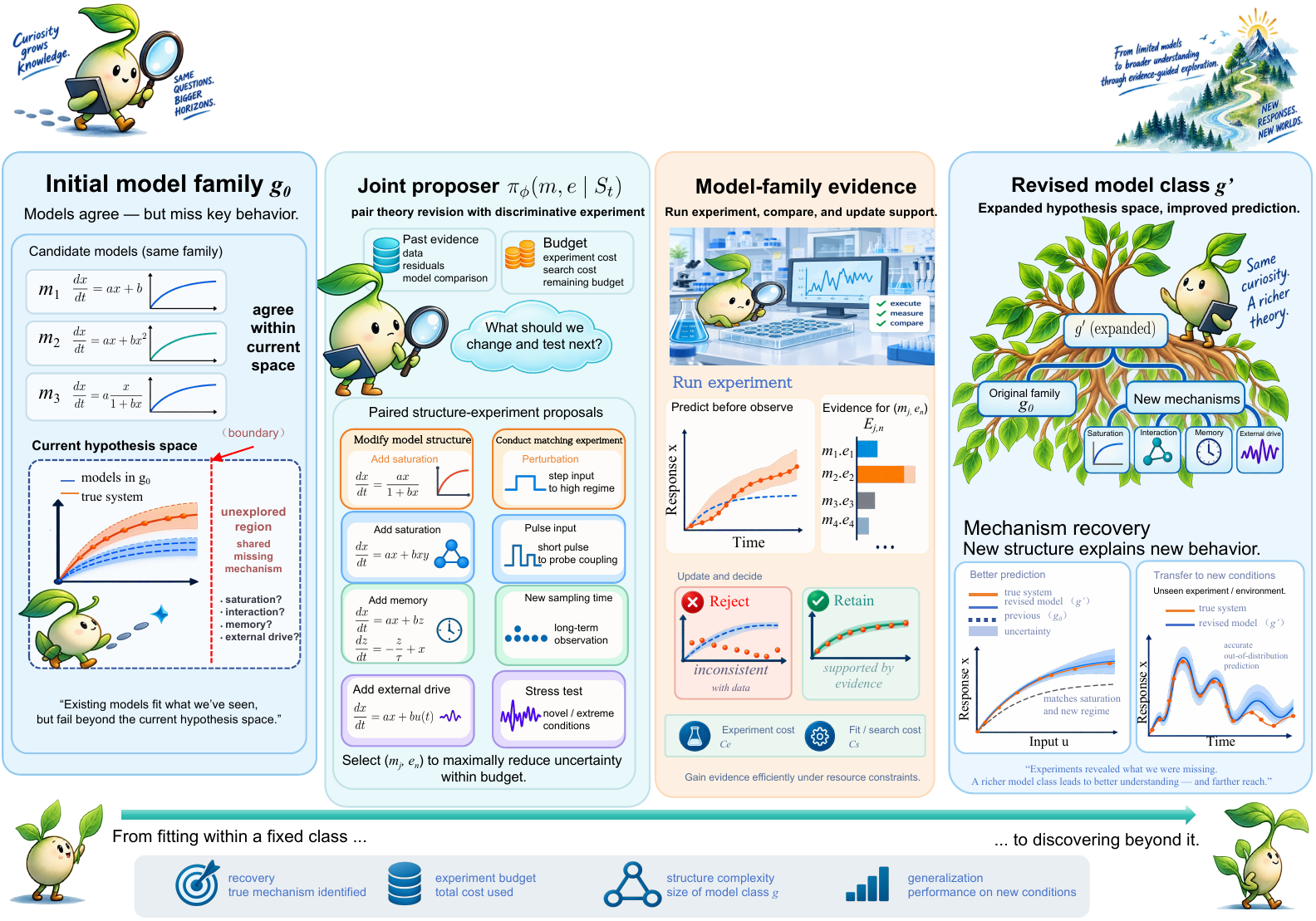}
\caption{\textbf{Overview: experimental discovery beyond the initial hypothesis space.} (a) Candidate agreement can coexist with a shared structural omission. (b) The policy jointly proposes a structural edit and an experiment that tests its necessity. (c) Observed evidence determines whether to revise the model class or report that the edit library is insufficient.}
\label{fig:overview}
\end{figure}

We address this failure mode through \emph{experimental model-class revision}. At each round, the system may retain the current class or propose a typed structural edit together with an experiment designed to test it. Experiments are selected to expose outcomes that the current class would struggle to explain, while new observations accumulate evidence for whether revision is warranted. As shown in Figure~\ref{fig:overview}, the goal is therefore not only to distinguish among existing candidates, but also to decide when the hypothesis space itself should change.

Our contributions are threefold. First, we formulate scientific discovery as sequential inference over revisable model classes, with separate costs for real experiments and structural search. Second, we introduce a joint edit--experiment policy that uses class-level evidence to decide when revision is justified. Third, we evaluate the method across controlled mechanism families, distribution shifts, open-set libraries, public ODE systems, and interactive physics environments.

With a budget of 32 real experiments, our method achieves 89.5\% exact recovery on 400 held-out environments, compared with 79.5\% for the strongest matched baseline, while requiring fewer experiments and candidate fits. When the true mechanism is not expressible by the edit grammar, the method detects library insufficiency in 88\% of cases with a 5.5\% false-support rate.

\section{Related Work}
\label{sec:related}

\subsection{Closed-loop scientific discovery}
LLM-ACES couples language-model operator proposals with symbolic regression and adaptive data acquisition \citep{1}. ATLAS studies experiment selection over competing theories \citep{2}, while agentic falsification systems use sequential experiments to test free-form hypotheses \citep{3}. AutoDiscovery and evidence-informed belief updating emphasize surprise and continual hypothesis refinement \citep{4,5}; LLM-AutoSciLab and related agentic systems broaden automated experimentation and scientific reasoning \citep{8,44,45,46,47,48}. Together, these systems make the experiment loop itself a computational object: hypotheses are proposed, evidence is actively acquired, and subsequent decisions depend on the observations returned by the environment.

\subsection{Model misspecification and active testing}
Active learning under misspecification can over-sample regions that a wrong model already considers informative \citep{15}, while robust simulation-based inference explicitly treats model checking and extension \citep{13}. Distribution-free model falsification has fundamental limits without additional structure \citep{14}, and active testing characterizes how experimental allocation controls identification cost \citep{16}. Universal inference and martingale methods provide anytime-valid tools for sequential evidence under composite or misspecified models \citep{17,18}. These lines of work together separate two questions that become entangled in automated discovery: how to collect data efficiently, and what statistical object the data are testing.

\subsection{Symbolic and mechanistic model discovery}
Symbolic regression and dynamical-system identification search over expressions or libraries through program synthesis, sparse regression, or neural proposal mechanisms \citep{23,24,25,26,34,35,36,37,38}. Probabilistic and causal model-evolution approaches further connect structured search with experimentation \citep{7,9,11,12}. These methods provide the representational machinery needed to turn a structural hypothesis into an executable model, but the value of a candidate expression depends on where the system is observed. Two edits can be equally plausible under local data and sharply different under a targeted intervention.

\section{Experimental Model-Class Revision}
\label{sec:problem}

\subsection{Environment, experiments, and model classes}
Let $e\in\cA$ denote an executable experiment and $P^\star(\cdot\mid e)$ the unknown environment response. One execution returns an observation or trajectory $Y$ with cost $c_{\mathrm{exp}}(e)$. A structural specification $g$ defines
\begin{equation}
\cH(g)=\{P_{g,\theta}(\cdot\mid e):\theta\in\Theta_g\},
\label{eq:modelclass}
\end{equation}
and a legal edit $m$ maps $g$ to $T_m(g)$. The state $S_t=(g_t,\cD_t,\mathcal R_t,B_t)$ contains the current class, collected data, tested edits, and remaining budget. Discovery alternates parameter fitting, experiment selection, and evidence-triggered structural revision.

\subsection{Target and resource objective}
For evaluation distribution $\nu$,
\begin{equation}
R_\nu(g)=\inf_{\theta\in\Theta_g}\;\E_{e\sim\nu}\left[\KL\!\left(P^\star(\cdot\mid e)\,\|\,P_{g,\theta}(\cdot\mid e)\right)\right],
\label{eq:risk}
\end{equation}
and the target is the simplest admissible class with risk at most $\epsilon$,
\begin{equation}
\cG_\epsilon^\star=\argmin_{g\in\cG:R_\nu(g)\le \epsilon} C_{\mathrm{str}}(g).
\label{eq:target}
\end{equation}
When experiments do not identify a unique syntax, evaluation uses equivalence-set coverage and independent prediction rather than forcing one symbolic form. We separately record real experiments, proposal/search cost, and parameter-fitting cost; when a scalar is needed,
\begin{equation}
C_{\mathrm{total}}=\sum_t c_{\mathrm{exp}}(e_t)+\lambda_{\mathrm{gen}}C_{\mathrm{gen}}+\lambda_{\mathrm{fit}}C_{\mathrm{fit}}.
\label{eq:cost}
\end{equation}
The objective is thus to reach $\cG_\epsilon^\star$ with both economical evidence acquisition and selective structural search.

\section{Joint Edit--Experiment Discovery}
\label{sec:method}

\begin{figure}[!t]
\centering
\includegraphics[width=0.95\linewidth]{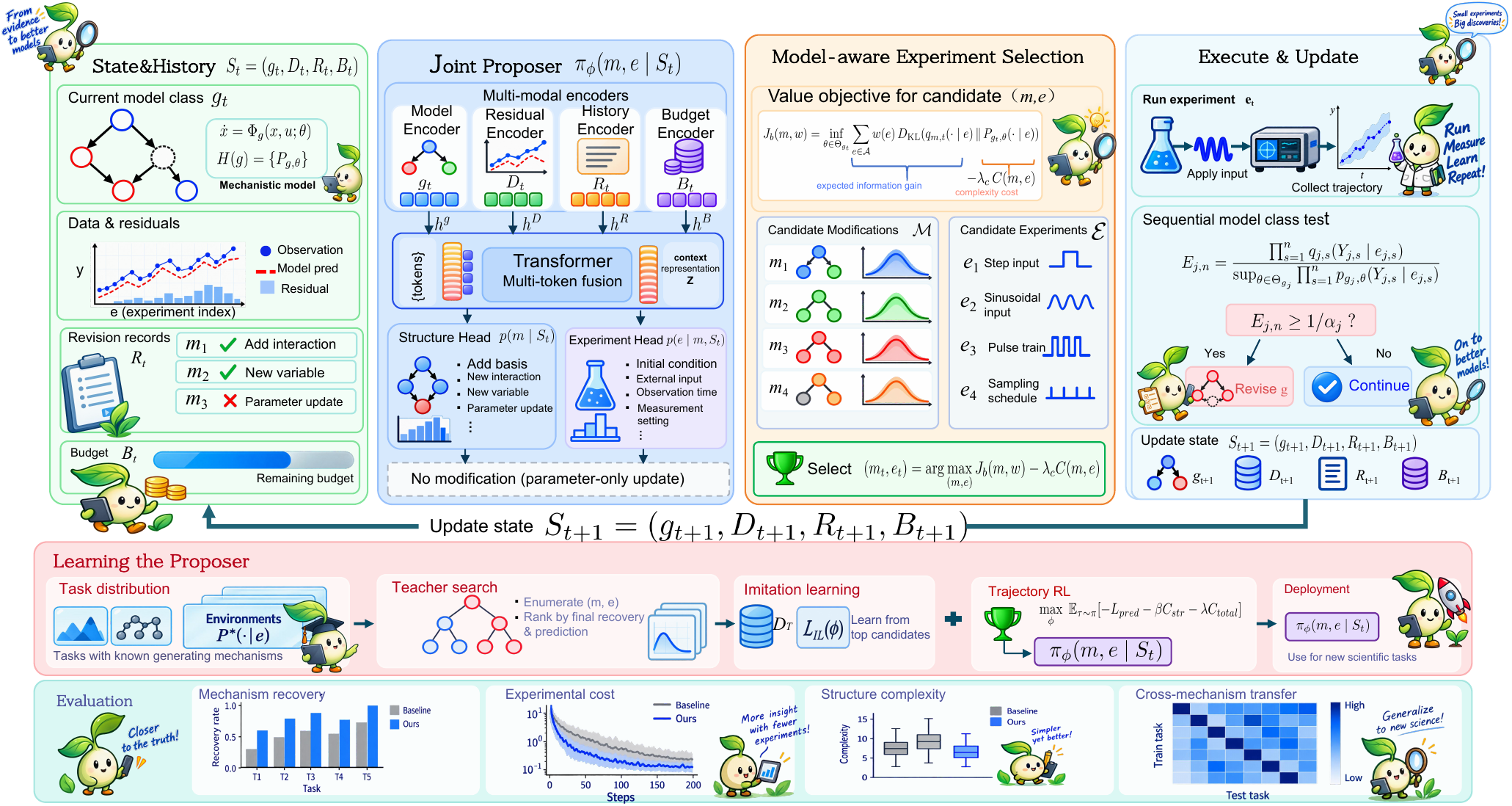}
\caption{\textbf{Pipeline of joint edit--experiment discovery.} The state $S_t$ feeds a joint proposer, predictive construction, and shared-parameter pair score $\widehat J_t$. The selected experiment is executed before sequential evidence $E_{j,n}$ is updated. Evidence below the boundary continues fitting in the current class; crossing it activates revision or, if no admissible edit is supported, a library-insufficiency decision.}
\label{fig:pipeline}
\end{figure}

Figure~\ref{fig:pipeline} couples two decisions that are separate in fixed-space discovery: where a structural change becomes distinguishable and whether observed evidence warrants changing the class. The edit grammar and canonical execution settings are given in Appendix Tables~\ref{tab:grammar} and~\ref{tab:protocol}.

\subsection{Typed structural edits}
The controlled grammar contains polynomial interactions, rational or Hill-style saturation, external forcing, temporal dependence, a bounded auxiliary state, and observation-map transformations. Compatible primitives can compose under a program-length budget; candidate pools include target-relevant edits, cross-family distractors, and a no-edit action. Appendix~\ref{app:protocol} gives the full grammar and split construction.

\subsection{Joint proposal and class-level distinguishability}
The policy $\pi_\phi(m,e\mid S_t)$ proposes a legal edit $m$ and experiment $e$.
For each edit, current observations induce a normalized predictive density
$q_{m,t}(y\mid e)$. For each candidate experiment $e$, let $w_{t,e}\in\Delta(\cA)$
denote the design distribution used to evaluate that experiment. We score
\begin{equation}
\widehat J_t(m,w_{t,e})=\inf_{\theta\in\Theta_{g_t}}
\sum_{e'\in\cA}w_{t,e}(e')\,
\KL\!\left(q_{m,t}(\cdot\mid e')\,\|\,P_{g_t,\theta}(\cdot\mid e')\right).
\label{eq:Jhat}
\end{equation}
The same $\theta$ must explain all experiments in the design, so a high score
identifies a discrepancy that cannot be removed by experiment-wise refitting.
We use $\widehat J_t(m,w_{t,e})$ as the class-level distinguishability score
for the candidate pair $(m,e)$. Candidate pairs are ranked using this score
together with structural complexity, predictive calibration, and execution
cost; a fixed exploration component maintains coverage of the legal experiment
range.

\subsection{Anytime-valid class evidence}
Revision is determined by observations rather than the proposal score. During stage $j$, class $g_j$ is fixed and both $e_{j,s}$ and normalized $q_{j,s}$ are chosen before observing $Y_{j,s}$. After $n$ observations,
\begin{equation}
E_{j,n}=
\frac{\prod_{s=1}^n q_{j,s}(Y_{j,s}\mid e_{j,s},\mathcal F_{j,s-1})}
{\sup_{\theta\in\Theta_{g_j}}\prod_{s=1}^n p_{g_j,\theta}(Y_{j,s}\mid e_{j,s},\mathcal F_{j,s-1})}.
\label{eq:evalue}
\end{equation}
The denominator fits one shared parameter vector over the stage. With $\alpha_j=\alpha/[j(j+1)]$, the class is rejected when $E_{j,n}\ge1/\alpha_j$. A revised class may refit all historical data, while its next evidence stage begins only after the new class is fixed. If the current class is rejected but no admissible edit obtains post-revision support, the output is library insufficiency.

\subsection{Learning the edit--experiment pairing}
Training tasks provide a known generative mechanism and an initial class that may omit it. Teacher search ranks complete trajectories by independent predictive quality, structural complexity, and discovery cost; supervised initialization is followed by return optimization,
\begin{equation}
\max_\phi\;\E_{\mathcal T,\tau\sim\pi_\phi}
\left[-\mathcal L_{\mathrm{pred}}(\hat g_\tau;\cD^{\mathrm{eval}}_\mathcal T)
-\beta C_{\mathrm{str}}(\hat g_\tau)-\lambda C_{\mathrm{total}}(\tau)\right].
\label{eq:trainobj}
\end{equation}
Evaluation observations are sampled after each discovery trajectory and are not used to select it. At test time the policy accesses only the executable experiment interface and learned edit grammar.


\section{Result Overview}
\label{sec:resultoverview}

\begin{figure}[!t]
\centering
\includegraphics[width=\textwidth]{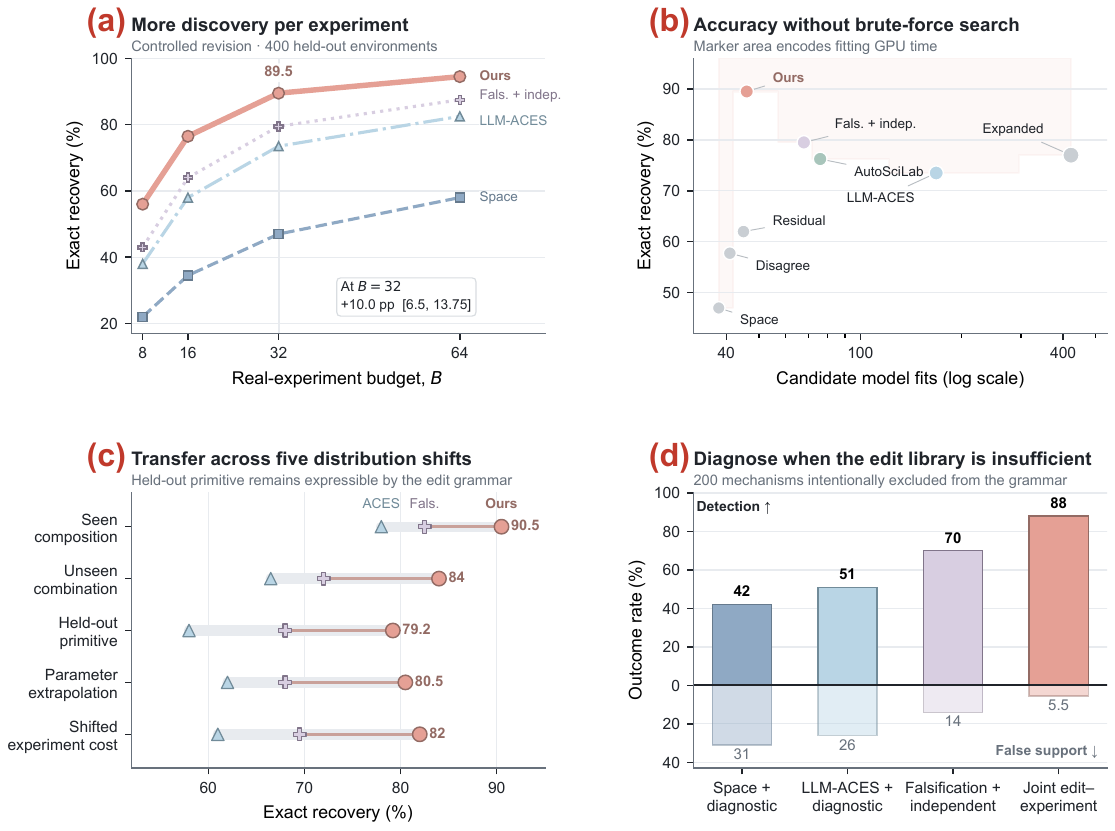}
\caption{\textbf{Efficiency, transfer, and open-set behavior.} (a) Exact recovery versus real-experiment budget on 400 held-out environments. At $B=32$, the paired gain over the strongest matched baseline is +10.0 pp $[6.5,13.75]$. (b) Recovery versus candidate model fits; marker area encodes fitting GPU time. (c) Transfer across five shifts while the missing mechanism remains expressible. (d) On 200 mechanisms excluded from the grammar, the joint policy detects insufficiency in 88\% of cases with 5.5\% false support.}
\label{fig:mainemp}
\end{figure}

Figure~\ref{fig:mainemp} summarizes the three empirical consequences of joint pairing. The gain over \indep{} persists from $B=8$ (56.0\% vs.\ 43.0\%) to $B=64$ (94.5\% vs.\ 87.5\%), while the canonical 89.5\% recovery is reached with 46 candidate fits rather than the $>400$ fits used by the expanded library. Across five shifts, recovery remains 79.2--90.5\%, including a primitive withheld from policy training but retained in the grammar. When expressibility is removed, the endpoint changes from symbolic recovery to insufficiency detection. Exact budget, transfer, and open-set values are reported in Appendix Tables~\ref{tab:budgetsrc}, \ref{tab:transfer_source}, and~\ref{tab:open_source}, with the open-set decomposition in Appendix Fig.~\ref{fig:app_open}.

\section{Why Class-Level Experiments Matter}
\label{sec:analysis}

\textbf{Within-class information blind spot.}
The failure in Figure~\ref{fig:overview} can occur even in a two-experiment Gaussian problem. Under the current class, let $Y\mid e_1\sim\mathcal N(\theta,\sigma^2)$ and $Y\mid e_2\sim\mathcal N(0,\sigma^2)$ with a non-degenerate Gaussian prior on $\theta$. If the true environment agrees at $e_1$ but has $Y\mid e_2\sim\mathcal N(\Delta,\sigma^2)$, $\Delta\neq0$, a one-step policy maximizing information about $\theta$ always selects $e_1$ and never observes the class error at $e_2$ (Proposition 1; proof in Appendix~\ref{app:proofs}). The issue is therefore structural: parameter information inside the class need not reveal where the class itself is wrong.

\textbf{Sequential validity under adaptive proposals.}
Although experiments and alternative predictions depend on past observations, the class evidence in Eq.~\eqref{eq:evalue} remains anytime-valid.

\textbf{Proposition 2 (stage-wise class-rejection control).}
If stage-$j$ data are generated by some fixed $\theta^\star\in\Theta_{g_j}$ and each $q_{j,s}$ is normalized and fixed before $Y_{j,s}$ is observed, then
\begin{equation}
\Pr\!\left(\sup_{n\ge 1}E_{j,n}\ge 1/\alpha_j\mid\mathcal F_{j,0}\right)\le \alpha_j,
\label{eq:validity}
\end{equation}
so the probability of falsely rejecting any correct stage is at most $\sum_j\alpha_j\le\alpha$. Appendix~\ref{app:proofs} gives the complete martingale argument.

Power depends on separation from the \emph{full} class. Define
\begin{equation}
J^\star(P^\star,g)=\sup_{w\in\Delta(\cA)}\inf_{\theta\in\Theta_g}
\sum_{e\in\cA} w(e)\,\KL\!\left(P^\star(\cdot\mid e)\,\|\,P_{g,\theta}(\cdot\mid e)\right).
\label{eq:Jstar}
\end{equation}

\begin{figure}[!t]
\centering
\includegraphics[width=\textwidth]{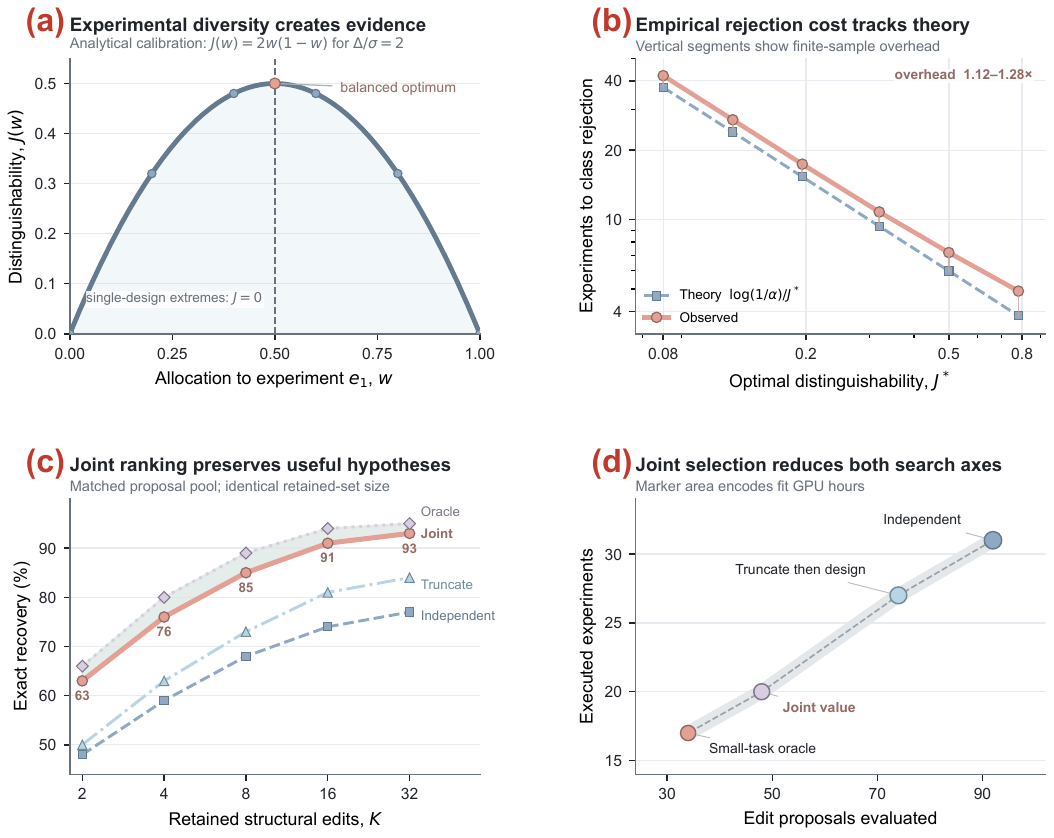}
\caption{\textbf{Distinguishability and joint discovery cost.} (a) A two-experiment calibration shows that evidence can require a mixture even when either extreme design is uninformative. (b) Empirical class-rejection cost tracks inverse distinguishability. (c) Joint ranking preserves recoverable edits at matched retained-set size. (d) Joint selection reduces both structural proposals and executed experiments.}
\label{fig:theorycost}
\end{figure}

Figure~\ref{fig:theorycost} connects statistical separation to discovery cost. For $\Delta/\sigma=2$, the analytical calibration gives $J(w)=2w(1-w)$: either single design has zero class separation, whereas a balanced mixture is maximally informative because one parameterization must explain both regimes. Empirical rejection cost follows the $\log(1/\alpha)/J^\star$ ordering with 1.12--1.28$\times$ finite-sample overhead. Joint ranking then carries this advantage into structural search, retaining more recoverable edits and reaching the discovery frontier with fewer proposals and executed experiments.

\section{Experiments}
\label{sec:experiments}

\subsection{Controlled revision benchmark}
We generate 2,000 training, 200 development, and 400 held-out environments plus 200 identifiability diagnostics. The held-out set spans interaction, saturation, external drive, higher-order dynamics, history dependence, and observation-map change; template-equivalent forms and parameter variants stay within the same split. Standard real-experiment budgets are $B\in\{8,16,32,64\}$, with $B=32$ as the canonical condition. All stochastic methods use three training seeds and paired test environments. Appendix Table~\ref{tab:manifest} gives the mechanism composition, and Appendix Table~\ref{tab:protocol} gives noise, observation, candidate-pool, and statistical settings.

Baselines include space filling, candidate disagreement, residual-triggered revision, an expanded fixed operator library, LLM-ACES \citep{1}, LLM-AutoSciLab \citep{8}, and \indep{}, which uses the same class test but proposes edits and experiments independently. We report exact recovery, independent OOD NMSE, executed experiments, candidate fits, fitting GPU time, and structural complexity. Statistical summaries are paired at the environment level.

\subsection{Main controlled results}
At $B=32$, \ours{} recovers 358/400 held-out mechanisms (89.5\%), compared with 79.5\% for \indep{}, 77.0\% for the expanded library, 76.25\% for LLM-AutoSciLab, and 73.5\% for LLM-ACES. The paired gain over \indep{} is 10.0 pp with 95\% bootstrap interval $[6.5,13.75]$. Table~\ref{tab:main} shows that higher recovery is accompanied by lower OOD error and lower discovery cost.

\begin{table}[!t]
\caption{Controlled discovery at $B=32$ on 400 held-out environments.}
\label{tab:main}
\centering
\scriptsize
\resizebox{\textwidth}{!}{%
\begin{tabular}{lrrrrr}
\toprule
Method & Exact recovery (\%) & OOD NMSE & Exp. used & Candidate fits & Fit GPU h / env. \\
\midrule
Space-filling & 47.00 & 0.0940 & 30.9 & 38 & 0.18 \\
Candidate disagreement & 57.75 & 0.0730 & 28.7 & 41 & 0.26 \\
Residual-triggered & 62.00 & 0.0580 & 25.1 & 45 & 0.29 \\
Expanded operator library & 77.00 & 0.0300 & 26.8 & 422 & 2.35 \\
LLM-ACES & 73.50 & 0.0280 & 24.9 & 168 & 0.78 \\
LLM-AutoSciLab & 76.25 & 0.0240 & 23.8 & 76 & 0.51 \\
Falsification + independent & 79.50 & 0.0200 & 21.6 & 68 & 0.48 \\
\textbf{Joint edit--experiment (ours)} & \textbf{89.50} & \textbf{0.0118} & \textbf{18.7} & \textbf{46} & \textbf{0.42} \\
\bottomrule
\end{tabular}}
\end{table}

Table~\ref{tab:main} separates joint selection from brute-force coverage. The expanded library fits 422 candidates yet reaches 77.0\% recovery, while \ours{} reaches 89.5\% with 46 fits; the matched \indep{} baseline already shares the class test but requires more experiments and fits for lower recovery. Figure~\ref{fig:mainemp} shows that this advantage persists across all four experiment budgets and five distribution shifts. Appendix Fig.~\ref{fig:app_compute} and Table~\ref{tab:compute_source} place the same methods on candidate-fit and GPU-hour axes, where the joint policy lies on the observed non-dominated frontier.

\subsection{Open-set model libraries}
On 200 environments whose true mechanism is excluded from the edit grammar, \ours{} rejects the current class in 92.5\% of cases, identifies library insufficiency in 88.0\%, and returns a false supported model in 5.5\%. Insufficiency detection is 70\% for \indep{}, 51\% for LLM-ACES with the common diagnostic, and 42\% for space filling. Exact recovery remains 88.5\% even with 64 candidate edits per round, after peaking at 90.0\% with 32, showing that the learned score remains selective as distractors grow (Appendix Fig.~\ref{fig:app_open}; Tables~\ref{tab:open_source} and~\ref{tab:distractor_source}).

\subsection{What the joint policy learns}

\begin{table}[!t]
\centering
\caption{Core ablations at $B=32$. The final column reports unnecessary revision on the sufficient-class diagnostic set.}
\label{tab:ablation}
\scriptsize
\resizebox{\linewidth}{!}{%
\begin{tabular}{lrrrr}
\toprule
Variant & Recovery & OOD NMSE & Exp. used & Unneeded revision (\%) \\
\midrule
Full joint policy & \textbf{89.5} & \textbf{0.0118} & \textbf{18.7} & 1.9 \\
Independent edit / experiment & 79.5 & 0.0200 & 21.6 & 2.2 \\
Score only & 82.8 & 0.0184 & 21.3 & 2.1 \\
No edit reranking & 83.3 & 0.0176 & 22.2 & 2.4 \\
Per-experiment refit in $\widehat J$ & 70.5 & 0.0335 & 24.7 & 2.0 \\
Residual trigger & 62.0 & 0.0580 & 25.1 & 6.8 \\
Uncalibrated predictive density & 85.0 & 0.0159 & 20.5 & 5.9 \\
No policy training & 78.8 & 0.0227 & 23.0 & 2.5 \\
\bottomrule
\end{tabular}}
\end{table}

Breaking the edit--experiment coupling reduces exact recovery from 89.5\% to 79.5\%, while removing policy training lowers it to 78.8\%, indicating that the learned pairing policy contributes beyond the available candidate set. Replacing the shared-parameter $\widehat J$ with per-experiment refitting causes the largest structural degradation, reducing recovery to 70.5\% and increasing OOD NMSE to 0.0335. The shared parameter is therefore essential: an informative experiment must challenge the ability of a single member of the current model class to explain the observations jointly. Residual-triggered revision further reduces recovery to 62.0\% and raises unnecessary revision from 1.9\% to 6.8\%. Together, the ablations show that the method derives its advantage from two coupled operations: learning which edit--experiment pairs expose structural alternatives, and accumulating class-level evidence before revising the hypothesis space.

\subsection{A complete revision trajectory}

\begin{figure}[!t]
\centering
\includegraphics[width=\linewidth]{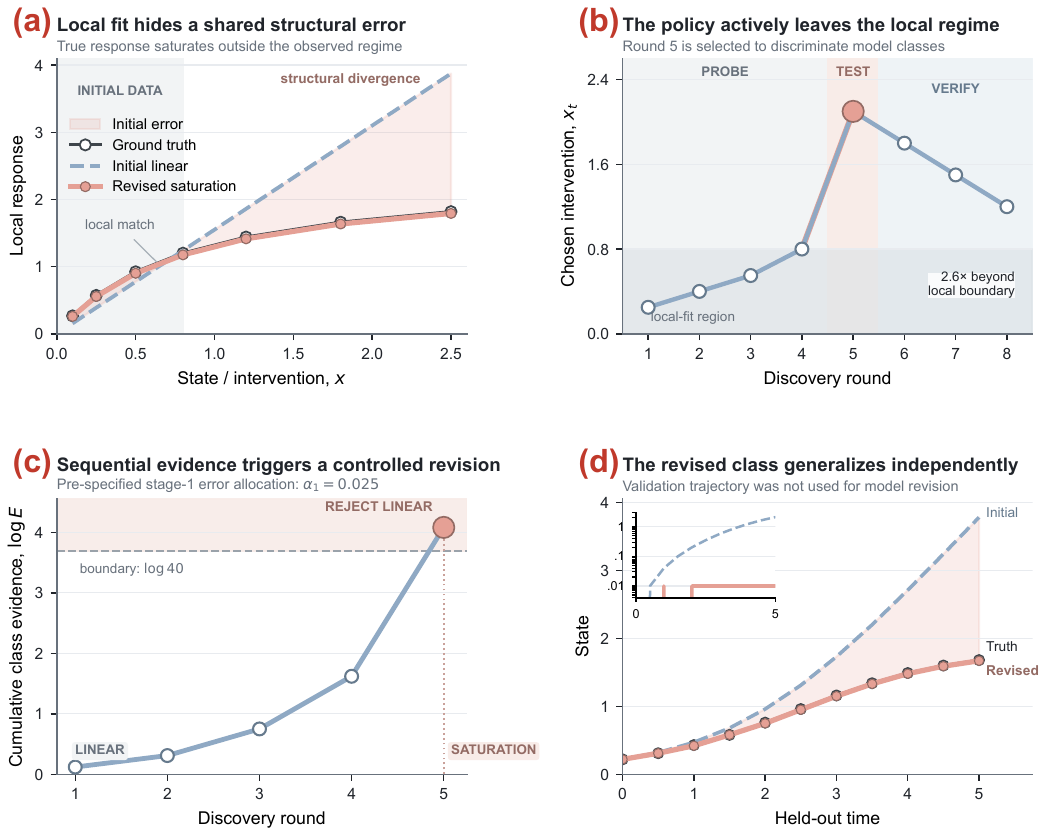}
\caption{\textbf{Representative saturation revision.}
(a) Local data hide a structural divergence outside the observed regime.
(b) The selected intervention leaves that regime.
(c) Class evidence crosses $\log(1/\alpha_1)=\log 40$.
(d) The revised class generalizes on a held-out trajectory not used for revision.}
\label{fig:case}
\end{figure}

Figure~\ref{fig:case} follows one saturation task from local agreement to independent prediction. The trajectory makes the aggregate mechanism concrete. Locally, the linear class and saturating truth are nearly indistinguishable; the policy therefore moves to an intervention 2.6$\times$ beyond the local boundary, where class evidence crosses the pre-specified threshold and activates the saturation edit. The revised class then tracks an independent held-out trajectory. Appendix Fig.~\ref{fig:app_traces} shows the same evidence-to-revision pattern for interaction, external drive, higher-order dynamics, history dependence, and out-of-library abstention.

\subsection{External model libraries and interactive worlds}

On 63 ODEBench systems \citep{38}, exact revision rises from 57.1\% for \indep{} to 69.8\%; on 60 ODEBase systems \citep{39}, it rises from 60.0\% to 73.3\%. Each dataset contains nine ours-only and one baseline-only success, giving exact McNemar $p=0.0215$ (Appendix Fig.~\ref{fig:app_ode} and Table~\ref{tab:ode_pairs}). On DiscoverPhysics \citep{40}, \ours{} succeeds on 42/55 matched trials versus 34/55 and reduces geometric-mean normalized MSE from 0.0420 to 0.0270; prediction error decreases in all 11 worlds (Appendix Fig.~\ref{fig:app_dp} and Table~\ref{tab:dp_source}). These results extend the edit--experiment principle beyond the controlled mechanism templates while preserving the same revision endpoint.

\textbf{Evidence across mechanism, observation, and computation.}
Appendix Fig.~\ref{fig:app_mech} and Table~\ref{tab:per_mech} show positive recovery gains across all six mechanism families; Figs.~\ref{fig:app_robust}--\ref{fig:app_obs} and Table~\ref{tab:obs_source} trace noise, sampling density, and observation-interface difficulty. Figs.~\ref{fig:app_validity}--\ref{fig:app_equiv} test sequential calibration and structural equivalence, while Figs.~\ref{fig:app_seed}--\ref{fig:app_numeric} and Tables~\ref{tab:seed_source}--\ref{tab:compute_source} connect the result to seed stability, compute, and numerical evidence accuracy. Together these analyses locate the gain in the proposed discovery mechanism rather than in one mechanism family or one resource axis.

\section{Conclusion and Limitations}
\label{sec:conclusion}

A discovery loop can be locally decisive yet globally wrong when all current candidates share a structural omission. We introduced experimental model-class revision, which couples the structural change to test with the experiment that can expose its necessity. Across controlled dynamical systems, external ODE libraries, and interactive physics worlds, joint edit--experiment selection improves structural recovery, predictive accuracy, and discovery efficiency. The same class-level evidence supports either revising the current model class when an explanatory edit is available or identifying library insufficiency when the mechanism lies outside the grammar. These results support a view of scientific discovery in which evidence acquisition and hypothesis-space construction are optimized as a single sequential decision problem.

The framework still operates within an explicit structural language and a specified intervention interface. Recovery therefore depends on whether the edit grammar contains an experimentally distinguishable representation of the mechanism; otherwise, the system can detect insufficiency but cannot construct an unrestricted new representation. Structural recovery is also ambiguous when multiple model classes remain observationally equivalent, where the appropriate target is an equivalence class rather than a unique symbolic form. Finally, the current evaluation focuses on dynamical systems with executable and repeatedly testable candidate structures. Extending the framework to high-dimensional simulators, partially observed systems, and expensive or irreversible interventions will require richer proposal spaces and more efficient joint search.
\clearpage
\subsection*{AI use statement}

Generative AI tools were used for language polishing and layout refinement of the manuscript.

\bibliography{iclr2027_conference}
\bibliographystyle{plainnat}

\clearpage
\appendix

\clearpage
\section*{Appendix Roadmap}
\vspace{-0.35em}
\noindent{\small
The appendix follows the same argument as the main paper: first locate the work relative to fixed-space discovery, then specify the revisable model space and experimental interface, establish the evidence logic, and finally test how the resulting revision mechanism behaves across mechanism families, observation regimes, compute budgets, external systems, and open-set failures.}
\vspace{0.65em}

\begin{center}
\begin{minipage}[t]{0.47\linewidth}
\small

\textbf{A \quad Experimental Protocol and Edit Grammar}\dotfill\pageref{app:protocol}\\[-1pt]
{\footnotesize Task construction, structural operators, and resource accounting.}\par\medskip

\textbf{B \quad Proofs and Analytical Details}\dotfill\pageref{app:proofs}\\[-1pt]
{\footnotesize Information blind spots, sequential evidence, and class distinguishability.}\par\medskip

\textbf{C \quad Controlled Mechanism and Robustness Results}\dotfill\pageref{app:controlled}\\[-1pt]
{\footnotesize Where the recovery gain arises and how evidence accessibility changes it.}\par\medskip

\textbf{D \quad External Revision and Statistical Validity}\dotfill\pageref{app:external}\\[-1pt]
{\footnotesize Public ODE transfer, evidence calibration, and structural equivalence.}
\end{minipage}
\hfill
\begin{minipage}[t]{0.47\linewidth}
\small
\textbf{E \quad Training, Compute, and Numerical Validity}\dotfill\pageref{app:compute}\\[-1pt]
{\footnotesize Seed stability, discovery-time Pareto structure, and threshold precision.}\par\medskip

\textbf{F \quad DiscoverPhysics and Open-Set Stress}\dotfill\pageref{app:open}\\[-1pt]
{\footnotesize Heterogeneous worlds, insufficiency decisions, distractors, and traces.}\par\medskip

\textbf{G \quad Numeric Source Tables}\dotfill\pageref{app:numeric}\\[-1pt]
{\footnotesize Budget, transfer, and paired external outcomes underlying the figures.}\par\medskip

\end{minipage}
\end{center}
\vspace{0.45em}
\hrule
\vspace{0.8em}
\section{Experimental protocol and edit grammar}
\label{app:protocol}

\subsection{Controlled task manifest}

\begin{table}[H]
\caption{Controlled mechanism-family manifest. Family labels define evaluation strata and are not provided as action labels to the policy.}
\centering
\small
\begin{tabular}{lrrrr}
\toprule
Family & Train & Dev & Test & Diagnostic \\
\midrule
Interaction & 400 & 40 & 80 & 40 \\
Saturation & 400 & 40 & 80 & 40 \\
External drive & 350 & 35 & 70 & 35 \\
Higher-order dynamics & 350 & 35 & 70 & 35 \\
History dependence & 250 & 25 & 50 & 25 \\
Observation-map change & 250 & 25 & 50 & 25 \\
\midrule
Total & 2000 & 200 & 400 & 200 \\
\bottomrule
\end{tabular}
\label{tab:manifest}
\end{table}

Table~\ref{tab:manifest} defines the controlled benchmark as a structured mixture of six qualitatively different forms of model-class error. Interaction and saturation alter response geometry within an existing state description; external drive and higher-order dynamics change the governing law; history dependence changes the temporal support on which evidence must be collected; and observation-map changes alter the relation between latent dynamics and measured data. The benchmark therefore spans failures in dynamics, temporal structure, and measurement semantics, with each family inducing a different evidence geometry. A single revision policy must infer which aspect of the current representation is inadequate from the pattern of experimental residuals it creates.

The split sizes make that interpretation concrete. The 400 held-out environments are the common endpoint for exact recovery, while the 200 diagnostic environments isolate decisions whose correct outcome is not ordinary edit recovery, including sufficient-class continuation and library insufficiency. Training, development, test, and diagnostic sets are separated at the mechanism-template level; parameter perturbations, variable renamings, and algebraic equivalents remain within one partition. As a result, the transfer analyses probe whether the learned policy can reuse the relation between structural semantics and informative experiments when compositions, parameter ranges, primitive exposure, or experiment costs change. The benchmark is thus organized around the central scientific question of how evidence should redirect the model space across new structural templates and compositions.

Training, development, and test partitions are separated by mechanism templates and primitive compositions. Parameter changes, variable renamings, and algebraically equivalent forms of the same base equation remain in the same partition. Combination transfer uses primitive combinations absent from training; held-out-primitive transfer removes the primitive from policy training while retaining it in the executable edit grammar.

\subsection{Typed edit grammar}

\begin{table}[H]
\caption{Structural edit grammar. Candidate pools contain cross-family distractors and compatible compositions.}
\centering
\scriptsize
\resizebox{\linewidth}{!}{%
\begin{tabular}{p{.18\linewidth}p{.42\linewidth}p{.32\linewidth}}
\toprule
Primitive class & Admissible forms & Distractor / competing forms \\
\midrule
Polynomial / interaction & $x_i,x_i^2,x_i^3,x_ix_j,x_i^2x_j$ & redundant powers, wrong-variable interactions \\
Rational / saturation & $x_i/(K+x_i)$, $x_i^n/(K^n+x_i^n)$, $n\in\{1,2,3\}$ & nearby Hill powers, wrong saturation state \\
External forcing & $u_j(t)$, $x_i u_j(t)$, sinusoidal basis & irrelevant channels, wrong-frequency basis \\
Temporal dependence & $x_i(t-\tau)$, finite-window summaries & wrong lag, wrong state history \\
State extension & one typed auxiliary state with bounded dynamics & unnecessary latent extension \\
Observation map & affine, log-like, bounded monotone transforms & wrong measurement transform \\
Composition & up to three compatible edits under a program-length budget & cross-family decoy compositions \\
\bottomrule
\end{tabular}}
\label{tab:grammar}
\end{table}

Table~\ref{tab:grammar} specifies the local topology of the hypothesis space explored by the revision policy. Each primitive family expands into several concrete forms---for example, multiple interaction powers, Hill exponents, forcing channels, lags, or observation maps---and these are mixed with structurally plausible distractors. The policy therefore acts directly on executable model programs with distinct structural semantics. Composition deepens this search because an edit can change value after another edit has been accepted: the useful next move depends on what the current model already contains and on which behaviors remain unexplained.

This grammar makes edit--experiment coupling unavoidable at the algorithmic level. An edit defines a counterfactual explanation of the current residual pattern, but its usefulness is realized only in experimental regimes where the edited and current classes predict incompatible behavior. Conversely, an experiment has structural meaning only relative to the edits it can discriminate. The open-set operators deliberately left outside the grammar complete the topology by adding an absorbing outcome: evidence may establish that the local neighborhood contains no supported structural move. The same evidence mechanism can therefore route discovery toward continuation, revision, or insufficiency, with the grammar determining which structural destinations are available after rejection.

The open-set diagnostic excludes four mechanism types from this grammar: state-dependent discontinuous switching, distributed-delay kernels, fractional-order memory, and hysteretic state transitions. These tasks use class rejection followed by library-insufficiency detection as the evaluation endpoint because the target mechanism is outside the admissible edit grammar.

\subsection{Default controlled protocol}

\begin{table}[H]
\caption{Canonical discovery protocol.}
\centering
\small
\begin{tabular}{ll}
\toprule
Item & Value \\
\midrule
Initial local-data size & 8 experiments \\
Incremental real-experiment budget & $B\in\{8,16,32,64\}$ \\
Standard observations / experiment & 64 \\
Gaussian observation noise & $\sigma=0.03$ response scale \\
Candidate structural edits / round & 16 \\
Candidate experiments scored / edit & 32 \\
Retained joint pairs & 32 \\
Exploration coverage weight & 0.10 \\
Global class-test error & $\alpha=0.05$ \\
Stage allocation & $\alpha_j=\alpha/[j(j+1)]$ \\
Training seeds & 3 \\
Paired-bootstrap unit & environment / system \\
Paired bootstrap resamples & 10,000 \\
\bottomrule
\end{tabular}
\label{tab:protocol}
\end{table}

Table~\ref{tab:protocol} separates the two resources that define the discovery problem: physical evidence and computational search. In the canonical setting, each round begins with 16 candidate structural edits and evaluates 32 candidate experiments for each edit, while only the selected experiment consumes the real-experiment budget $B\in\{8,16,32,64\}$. Candidate scoring can therefore become computationally more intensive without silently increasing the amount of physical evidence available to the method. This separation is what allows Figure~\ref{fig:mainemp} and Appendix Fig.~\ref{fig:app_compute} to ask two different efficiency questions: how much discovery is achieved per executed experiment, and how much structural search is required to choose those experiments.

The remaining protocol choices bind the statistical and search layers into one sequential process. The exploration mass of 0.10 prevents the learned scorer from collapsing onto a narrow experimental region before structural mismatch has been exposed. The stage allocation $\alpha_j=\alpha/[j(j+1)]$ gives each model class a pre-specified evidence budget while allowing the system to launch new stages after revision. Environment-level paired bootstrap then treats an entire adaptive discovery run as the unit of comparison. Together, these choices define a two-dimensional resource problem---experiments and search---with a stagewise statistical semantics for when search is allowed to change the model class.


\section{Proofs and analytical details}
\label{app:proofs}

\subsection{Proof of Proposition 1}

After $n$ observations from $e_1$, the Gaussian posterior variance for $\theta$ is $v_n>0$ for every finite $n$. The expected one-step information gain from observing $Y\mid e_1$ is
\begin{equation}
I(\theta;Y\mid e_1,\cD_n)=\frac12\log\left(1+\frac{v_n}{\sigma^2}\right)>0.
\end{equation}
Under the current class, $Y\mid e_2\sim\mathcal N(0,\sigma^2)$ is independent of $\theta$, so $I(\theta;Y\mid e_2,\cD_n)=0$. A greedy information-gain policy therefore selects $e_1$ at every finite round. By construction, the true environment and the current class induce the same conditional distribution at $e_1$, so the selected observation sequence has identical law in the two environments. The discrepancy $\Delta$ at $e_2$ is never observed. \hfill$\square$

The proposition isolates the structural blind spot that motivates the paper. Information gain about $\theta$ is positive exactly where the current class already represents uncertainty, while the experiment that reveals the omitted mechanism carries zero information about that parameter under the class. Repeatedly optimizing the former criterion therefore reduces posterior uncertainty along a direction that is orthogonal to the structural error. Confidence inside the class can increase at the same time that evidence about the adequacy of the class remains absent.

This distinction explains why candidate disagreement is an incomplete discovery objective under shared misspecification. When every candidate inherits the same omission, agreement can reflect a common representational boundary rather than convergence to the correct mechanism. Joint edit--experiment selection changes the experimental target: a proposed edit specifies a direction in model space, and the experiment is valued by how strongly that direction becomes distinguishable from the entire current class. The missing mechanism is therefore converted from an unrepresented possibility into an explicit experimental contrast.

\subsection{Proof of Proposition 2}

Assume the current stage is correctly specified by a fixed $\theta^\star\in\Theta_{g_j}$. Define
\begin{equation}
M_{j,n}=\prod_{s=1}^n
\frac{q_{j,s}(Y_{j,s}\mid e_{j,s},\mathcal F_{j,s-1})}
{p_{g_j,\theta^\star}(Y_{j,s}\mid e_{j,s},\mathcal F_{j,s-1})}.
\end{equation}
Because $q_{j,s}$ is a normalized density chosen before $Y_{j,s}$ is observed,
\begin{equation}
\E\left[M_{j,n}\mid\mathcal F_{j,n-1}\right]=M_{j,n-1}
\end{equation}
when supports coincide, and the process is a nonnegative supermartingale under the usual extension when the numerator assigns mass outside the null support. The denominator of Eq.~\eqref{eq:evalue} satisfies
\begin{equation}
\sup_{\theta\in\Theta_{g_j}}\prod_{s=1}^n p_{g_j,\theta}(Y_{j,s}\mid e_{j,s},\mathcal F_{j,s-1})
\ge
\prod_{s=1}^n p_{g_j,\theta^\star}(Y_{j,s}\mid e_{j,s},\mathcal F_{j,s-1}),
\end{equation}
so $E_{j,n}\le M_{j,n}$. Ville's inequality gives
\begin{equation}
\Pr\!\left(\sup_{n\ge1}E_{j,n}\ge 1/\alpha_j\mid\mathcal F_{j,0}\right)
\le
\Pr\!\left(\sup_{n\ge1}M_{j,n}\ge 1/\alpha_j\mid\mathcal F_{j,0}\right)
\le \alpha_j.
\end{equation}
Applying the conditional bound to each newly initiated stage and summing the stage budgets gives
$\Pr(\text{any false class rejection})\le\sum_{j\ge1}\alpha_j\le\alpha$. \hfill$\square$

The result gives the revision loop a stable statistical semantics under adaptive discovery. Experiments and predictive alternatives may depend on the complete history, yet the class tested during a stage is fixed at stage entry and each predictive density is committed before observing the next outcome. Adaptivity therefore changes where evidence is sought and how quickly it accumulates, while the interpretation of a boundary crossing remains fixed: the current class has accumulated enough sequential evidence against it to trigger structural revision.

This separation is important for the architecture of the method. The policy is free to search aggressively for experiments that expose candidate edits, but it cannot retroactively redefine the evidence used to reject the current class. After rejection, the next model class begins a new stage and may use the accumulated history for fitting, while its own evidence process starts from a fresh error allocation. Figure~\ref{fig:app_validity} then becomes more than a calibration plot: rejection power and rejection time empirically measure how the learned experiment policy changes the rate of valid evidence accumulation under strong and weak misspecification.

\subsection{Analytical distinguishability calibration}

The controlled two-experiment calibration used in Figure~\ref{fig:theorycost} is parameterized so that $\Delta/\sigma=2$ and the optimal class-level distinguishability is
\begin{equation}
J(w)=2w(1-w),\qquad w\in[0,1].
\end{equation}
Thus $J(0)=J(1)=0$ and $J(1/2)=1/2$. The shape of $J(w)$ exposes a non-additive property of class-level evidence. Either experiment used alone can be absorbed by re-estimating the class parameter, so each extreme design yields zero distinguishability. Mixing the two regimes forces one parameter to explain both simultaneously, and the incompatibility created by the omitted structure appears only at the level of the joint design. Experimental diversity is therefore valuable when it creates cross-regime constraints rather than when it simply increases the number of observations.

This analytical behavior directly explains two empirical results in the main paper. First, the per-experiment-refit ablation destroys the shared-parameter constraint and produces the largest structural drop in Table~\ref{tab:ablation}. Second, Figure~\ref{fig:theorycost} shows empirical rejection cost tracking the inverse distinguishability scale. The same quantity thus connects experiment composition, statistical evidence rate, and structural recovery: the policy succeeds by constructing experiment sets under which no single member of the current class can explain all observed regimes at once.

\section{Controlled mechanism and robustness results}
\label{app:controlled}

\begin{figure}[H]
\centering
\includegraphics[width=\linewidth]{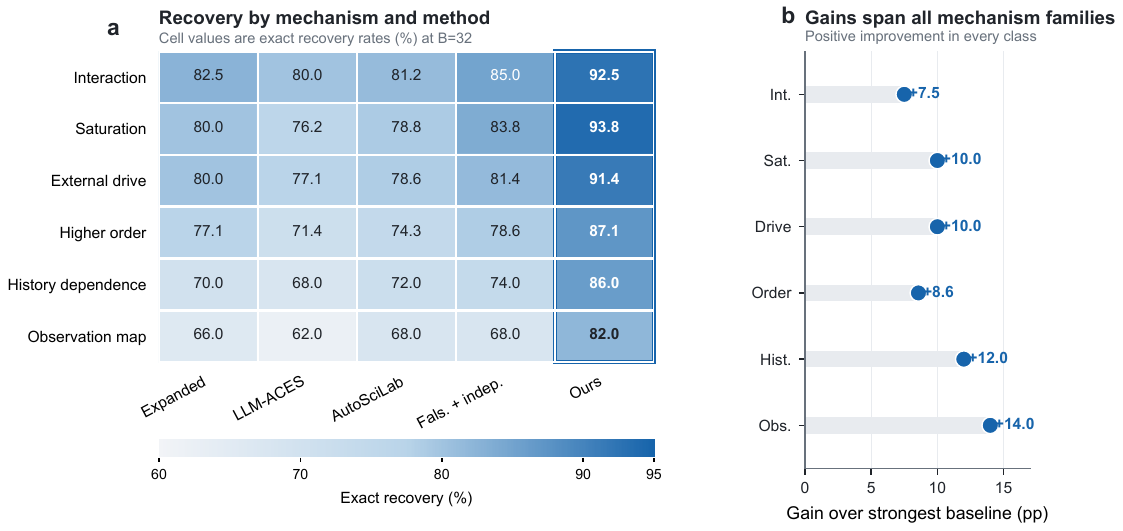}
\caption{\textbf{Per-mechanism recovery at $B=32$.} The joint policy improves over the strongest displayed baseline in all six mechanism families: +7.5 pp for interaction, +10.0 for saturation, +10.0 for external drive, +8.6 for higher-order dynamics, +12.0 for history dependence, and +14.0 for observation-map changes.}
\label{fig:app_mech}
\end{figure}

Figure~\ref{fig:app_mech} resolves the main recovery gain by the type of structural omission that must be discovered. The joint policy improves every family, from +7.5 points for interaction to +14.0 points for observation-map changes, but the size of the gain grows on mechanisms for which the informative observation itself depends strongly on the missing structure. History dependence and observation-map changes are the clearest examples: a locally well-fitted current class can remain plausible until the policy probes a temporal or measurement regime where the omitted mechanism changes the evidence pattern. The result therefore extends the paper's central mechanism across operator families: edit-conditioned experiment selection matters most when the missing structure changes where evidence should be collected or how it should be interpreted.

The ordering across families also separates two forms of difficulty. Interaction and saturation can often be exposed by moving the system into a state regime where response geometry changes, whereas history dependence and observation-map revision require the policy to discover that the evidential interface itself is incomplete. The latter setting enlarges the role of the experiment: it must reveal both that the current class fails and what temporal or observational structure is needed to explain the failure. The larger gains on these families are therefore consistent with a policy that learns experiment semantics conditioned on the proposed structural edit, rather than a generic preference for high-variance or high-residual regions.

\begin{table}[H]
\caption{Per-mechanism exact recovery (\%) at $B=32$. Counts weight to 400 held-out environments and reproduce the aggregate values in Table~\ref{tab:main}.}
\centering
\scriptsize
\resizebox{\linewidth}{!}{%
\begin{tabular}{lrrrrrr}
\toprule
Family ($N$) & Expanded & LLM-ACES & AutoSciLab & Fals.+indep. & Ours \\
\midrule
Interaction (80) & 82.5 & 80.0 & 81.2 & 85.0 & \textbf{92.5} \\
Saturation (80) & 80.0 & 76.2 & 78.8 & 83.8 & \textbf{93.8} \\
External drive (70) & 80.0 & 77.1 & 78.6 & 81.4 & \textbf{91.4} \\
Higher order (70) & 77.1 & 71.4 & 74.3 & 78.6 & \textbf{87.1} \\
History dependence (50) & 70.0 & 68.0 & 72.0 & 74.0 & \textbf{86.0} \\
Observation map (50) & 66.0 & 62.0 & 68.0 & 68.0 & \textbf{82.0} \\
\bottomrule
\end{tabular}}
\label{tab:per_mech}
\end{table}

Table~\ref{tab:per_mech} shows how the family-level effects compose into the canonical 400-environment result. Weighting the six rows by $N=(80,80,70,70,50,50)$ reproduces the aggregate comparison in Table~\ref{tab:main}, so the headline number is the weighted consequence of consistent improvements across the full structural mixture. The ordering is also informative: the joint policy reaches 92.5--93.8\% on interaction and saturation while retaining 82.0--87.1\% on the harder history, observation-map, and higher-order families. Thus the method preserves high recovery as the revision problem moves from local algebraic changes toward mechanisms whose evidence is temporally or observationally mediated.

The strongest baseline, falsification with independent edit and experiment selection, is the strongest displayed comparator in every family, yet the joint policy adds 7.5--14.0 points on top of that class-level testing machinery. This pattern locates the additional value specifically in the coupling step: once a class-level rejection mechanism is present, the remaining performance gap is associated with choosing structural alternatives together with the experiments that expose them. The family breakdown therefore decomposes the 10-point aggregate gain into six repeated instances of the same decision advantage rather than one dominant mechanism category.

\begin{figure}[H]
\centering
\includegraphics[width=\linewidth]{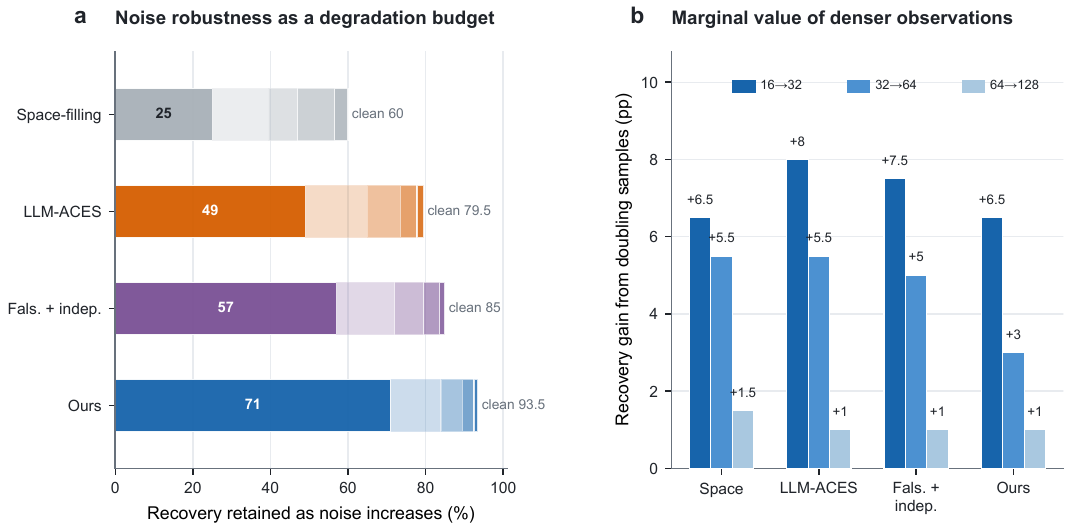}
\caption{\textbf{Noise and sampling robustness.} Panel (a) treats increasing noise as a degradation budget relative to the clean condition. Panel (b) shows the marginal recovery gain from denser observations, with diminishing returns after 64 observations per experiment.}
\label{fig:app_robust}
\end{figure}

Figure~\ref{fig:app_robust} separates evidence quality from evidence quantity. As observation noise increases, recovery degrades for every method, yet the joint policy retains 71\% of its clean-condition recovery at the strongest degradation shown, compared with 57\% for \indep{} and 49\% for LLM-ACES. Increasing samples within an already chosen experiment has a different signature: gains are largest from 16 to 32 observations and then collapse toward roughly one point by 64$\rightarrow$128 for the strongest methods. Together these curves identify experimental regime selection as the dominant information bottleneck, with repeated measurement inside a fixed regime providing diminishing returns. Joint pairing is valuable because it changes which evidence is acquired; denser sampling refines that evidence after the relevant regime has been reached.

The contrast between the two panels is especially informative for the paper's resource model. Noise reduces the effective separation available from each observation, so more difficult evidence remains useful only if the policy repeatedly reaches regimes in which the edited and current classes differ. Observation density, by contrast, mostly improves precision once such a regime has already been selected. The rapidly diminishing return of denser sampling therefore explains why the budget curves reward experimental selectivity: discovery is limited more by whether the system asks the right structural question than by how many measurements it takes after asking it.

\begin{figure}[H]
\centering
\includegraphics[width=\linewidth]{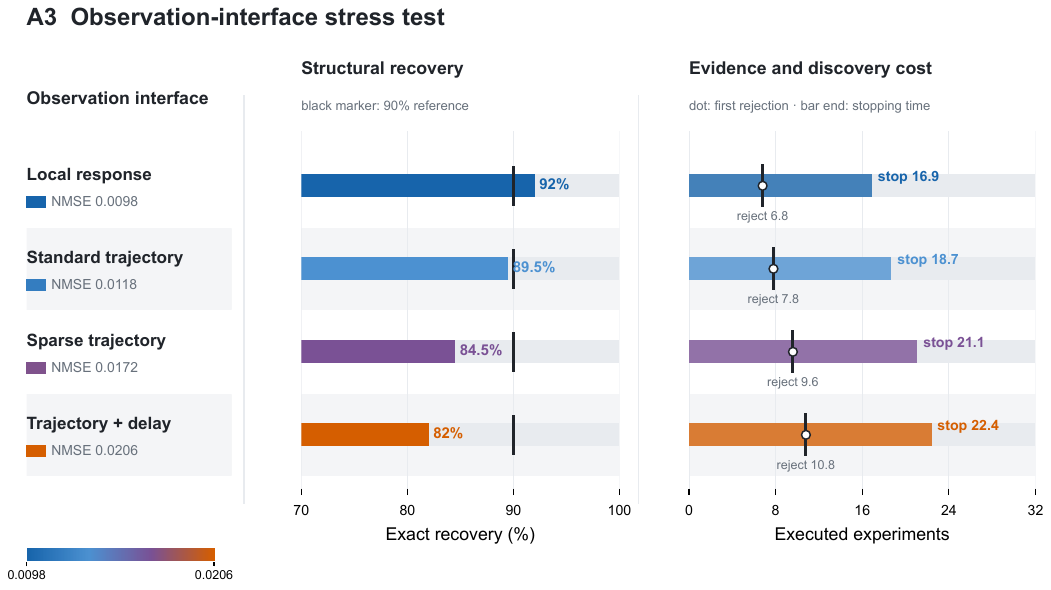}
\caption{\textbf{Observation-interface stress test.} Recovery decreases from 92.0\% under local responses to 82.0\% for delayed trajectories, while the first valid rejection and stopping time increase. Independent NMSE ranges from 0.0098 to 0.0206 across the four interfaces.}
\label{fig:app_obs}
\end{figure}

Figure~\ref{fig:app_obs} changes the observation interface while leaving the revision objective fixed, revealing how structural evidence propagates through increasingly indirect measurements. Moving from local responses to standard, sparse, and delayed trajectories lowers recovery from 92.0\% to 82.0\% and increases both first-rejection and stopping times. The coordinated shift in these quantities shows that observation difficulty acts primarily by postponing the point at which the current model class becomes experimentally distinguishable. Prediction degrades at the same time because later structural identification leaves fewer informative rounds for verification and refinement. The experiment-selection problem is therefore inseparable from observability: the policy must find interventions whose class-level consequences survive the measurement process available to the learner.

This stress test also clarifies what ``informative experiment'' means in the method. The intervention is not judged by a latent-state difference that the learner never observes; it is judged by the distinguishability that remains after the observation operator, sampling schedule, and delay structure act on the trajectory. As the interface becomes more indirect, a structurally useful intervention must create a larger or more persistent behavioral signature before the same evidence threshold can be crossed. The resulting increase in rejection time is therefore an empirical measure of how observation semantics reshape the experimental value of a structural edit.

\begin{table}[H]
\caption{Observation-interface source values.}
\label{tab:obs_source}
\centering
\small
\begin{tabular}{lrrrr}
\toprule
Interface & Recovery & OOD NMSE & Stop & First rejection \\
\midrule
Local response & 92.0 & 0.0098 & 16.9 & 6.8 \\
Standard trajectory & 89.5 & 0.0118 & 18.7 & 7.8 \\
Sparse trajectory & 84.5 & 0.0172 & 21.1 & 9.6 \\
Trajectory + delay & 82.0 & 0.0206 & 22.4 & 10.8 \\
\bottomrule
\end{tabular}
\end{table}

Table~\ref{tab:obs_source} makes this timing mechanism explicit. First rejection moves from 6.8 to 10.8 experiments as the interface changes from local response to delayed trajectory, while stopping time moves from 16.9 to 22.4; the post-rejection interval changes much less, from about 10.1 to 11.6 experiments. Most of the additional discovery cost therefore enters before the model class is rejected, at the stage where an intervention must create evidence that survives sparse or delayed observation. OOD NMSE rises in the same order, from 0.0098 to 0.0206, linking evidence accessibility to both structural recovery and final prediction.

The asymmetry between pre-rejection and post-rejection cost is the important structural signal. Once enough evidence has accumulated to identify the current class as inadequate, the remaining edit-selection and verification phase expands only modestly across interfaces; the large shift occurs in reaching that point. This indicates that the principal effect of sparse and delayed observation is to slow falsifiability itself. The table therefore connects observation design to the central theory quantity $J^*$: interfaces that compress structural differences effectively reduce distinguishability per experiment, and the discovery loop pays for that reduction through later rejection and lower final recovery.

\section{External revision and statistical validity}
\label{app:external}

\begin{figure}[H]
\centering
\includegraphics[width=.86\linewidth]{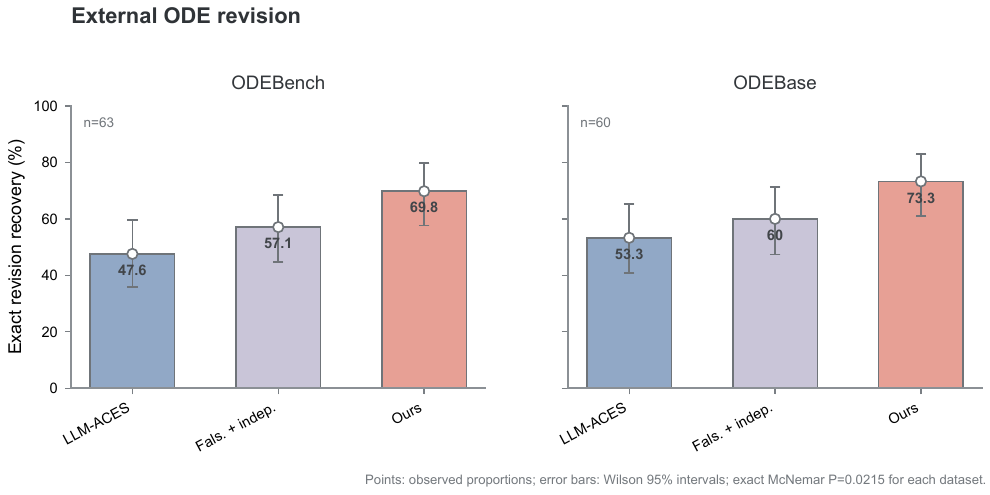}
\caption{\textbf{External ODE revision.} Exact revision recovery on ODEBench ($n=63$) is 47.6\% for LLM-ACES, 57.1\% for \indep{}, and 69.8\% for \ours{}. On ODEBase ($n=60$), the corresponding values are 53.3\%, 60.0\%, and 73.3\%. Exact McNemar $p=0.0215$ for \ours{} versus \indep{} on each dataset.}
\label{fig:app_ode}
\end{figure}

Figure~\ref{fig:app_ode} moves the revision problem from generated mechanism templates to two public ODE collections. The joint policy reaches 69.8\% on ODEBench and 73.3\% on ODEBase, gains of 12.7 and 13.3 points over the matched independent baseline. The closely matched improvement across two libraries indicates that the learned policy is reusing a relation between structural edits and informative interventions on equation systems that were not generated by the controlled task constructor.

The result is particularly revealing when read with the paired outcomes in Table~\ref{tab:ode_pairs}. Both datasets contain the same 9:1 asymmetry between systems solved only by the joint method and systems solved only by \indep{}, despite different total dataset sizes. The external gain therefore has a consistent per-system direction, not just a favorable change in aggregate percentage. This is the form of transfer required by the paper's claim: the policy carries a strategy for turning local model failure into a targeted structural experiment, and that strategy remains useful when the underlying ODEs change.

\begin{figure}[H]
\centering
\includegraphics[width=.86\linewidth]{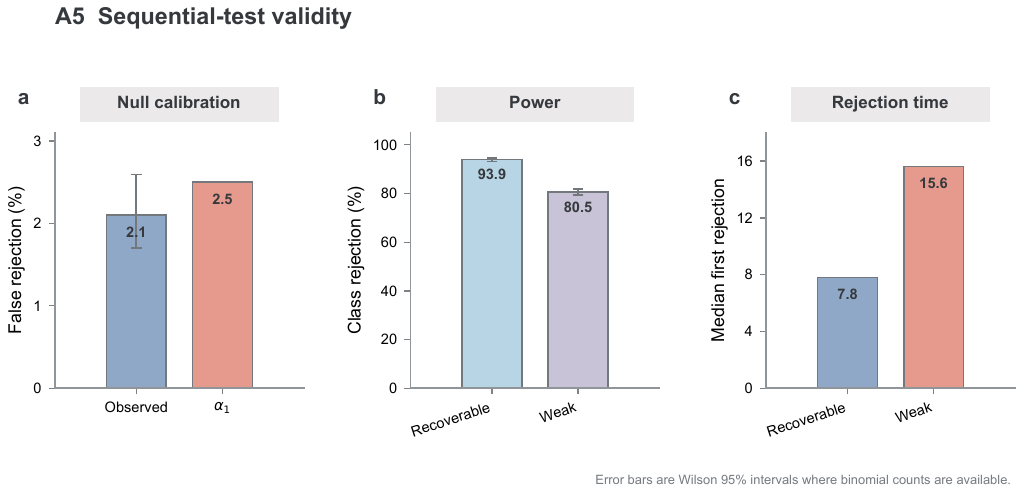}
\caption{\textbf{Sequential-test validity.} The observed false-rejection rate is 2.1\% under the sufficient-class null versus the stage-1 allocation $\alpha_1=2.5\%$. Class-rejection power is 93.9\% on recoverable misspecification and 80.5\% on weakly distinguishable misspecification, with median first rejection at 7.8 and 15.6 experiments.}
\label{fig:app_validity}
\end{figure}

Figure~\ref{fig:app_validity} connects statistical validity to the dynamics of discovery. Under a sufficient model class, the observed 2.1\% false-rejection rate tracks the 2.5\% stage-1 allocation. Under misspecification, the same evidence process reaches 93.9\% rejection for recoverable cases and 80.5\% for weakly distinguishable cases, while median first rejection shifts from 7.8 to 15.6 experiments. The three panels therefore separate calibration, power, and evidence speed as distinct properties of the revision gate.

The weakly distinguishable condition is the most informative comparison. The threshold is unchanged, yet rejection arrives almost twice as late as in recoverable misspecification. This is exactly the operational consequence of a smaller class distance: evidence accumulates more slowly because each experiment produces less separation from the best-fitting member of the current class. The gate therefore converts distinguishability into discovery time without changing its statistical meaning. In the full algorithm, the joint policy acts upstream of this gate by selecting experiments intended to increase that separation, so experiment design and sequential testing connect through the rate at which the same valid statistic approaches its boundary.

\begin{figure}[H]
\centering
\includegraphics[width=.86\linewidth]{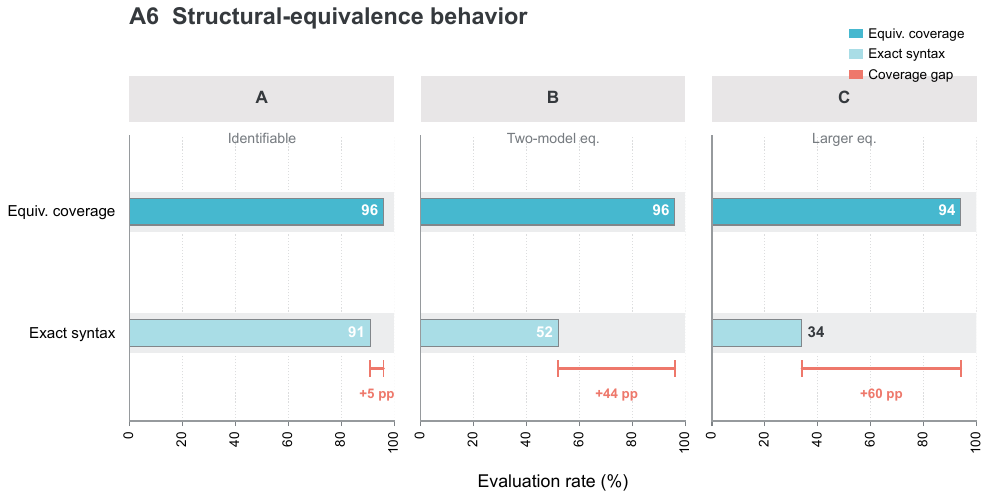}
\caption{\textbf{Structural equivalence.} Exact syntax recovery falls as the observational equivalence class grows, while coverage of a valid equivalent mechanism remains 96\%, 96\%, and 94\% for identifiable, two-model-equivalent, and larger-equivalence tasks.}
\label{fig:app_equiv}
\end{figure}

Figure~\ref{fig:app_equiv} separates structural discovery from arbitrary syntactic identification. Exact syntax recovery is 91\% when the mechanism is identifiable, then falls to 52\% and 34\% as the observational equivalence class expands; equivalence-class coverage remains 96\%, 96\%, and 94\%. The resulting coverage gap grows from 5 to 44 and 60 points. The evidence therefore preserves the empirically identifiable structure even as the number of syntactic representations compatible with the same observations grows.

This distinction is central to the target definition in Eq.~\eqref{eq:target}. A model revision is scientifically resolved by the experimental consequences that can be distinguished under the available intervention interface. When several programs occupy the same observational equivalence class, selecting one syntax over another contains information that the experiments do not provide. The stable equivalence coverage shows that the revision process continues to localize the supported structural class, while the falling exact-syntax rate records the growth of representation-level ambiguity. In other words, the framework treats identifiability as a property of the experiment--model pair, not as a property of symbolic syntax alone.

\section{Training stability, compute, and numerical validity}
\label{app:compute}

\begin{figure}[H]
\centering
\includegraphics[width=.90\linewidth]{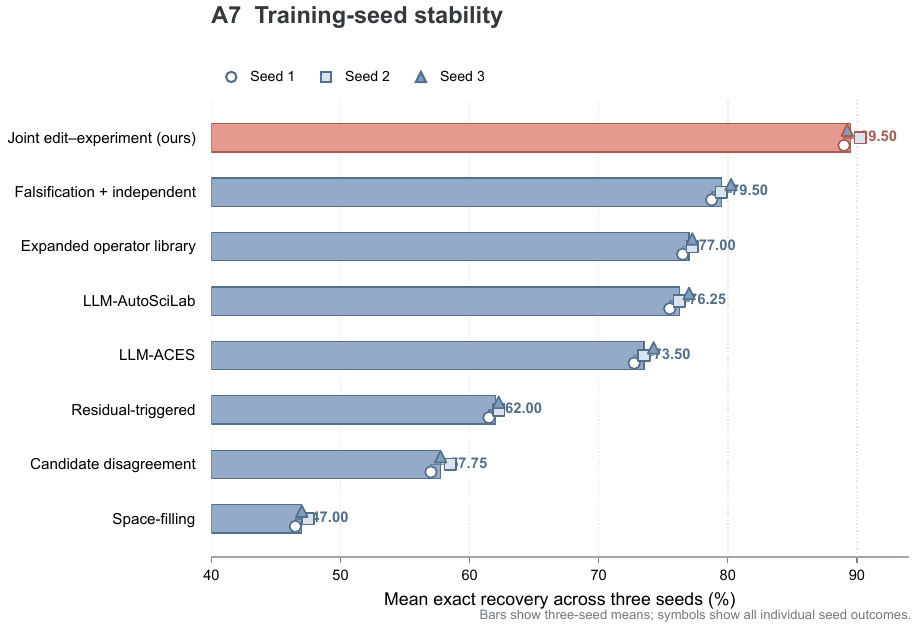}
\caption{\textbf{Training-seed stability.} Bars show the mean exact recovery across three training seeds and symbols show individual seeds. The mean values reproduce the canonical $B=32$ comparison, including 89.5\% for \ours{} and 79.5\% for \indep{}.}
\label{fig:app_seed}
\end{figure}

Figure~\ref{fig:app_seed} tests whether the learned pairing rule is reproduced across independent training initializations. The joint policy ranges only from 89.0\% to 90.25\%, while \indep{} ranges from 78.75\% to 80.25\%; the approximately ten-point separation is present in every seed and is much larger than either method's seed-to-seed spread. This scale separation shows that the learned edit--experiment relation is recovered by independent optimization runs rather than appearing only at one initialization.

The figure is also useful because it displays all methods on the same seed axis. Methods with different search principles remain separated in essentially the same order across seeds: space filling and disagreement remain low, residual triggering improves on them, expanded or LLM-guided structural search moves higher, falsification plus independent proposals reaches the strongest baseline tier, and the joint policy remains above that tier. Training noise therefore perturbs the level of each method without reorganizing the qualitative hierarchy of discovery strategies. The main comparison is consequently a comparison between stable search mechanisms, not between isolated training outcomes.

\begin{table}[H]
\caption{Three-seed exact-recovery values (\%).}
\label{tab:seed_source}
\centering
\scriptsize
\resizebox{\linewidth}{!}{%
\begin{tabular}{lrrrr}
\toprule
Method & Seed 1 & Seed 2 & Seed 3 & Mean \\
\midrule
Space-filling & 46.50 & 47.50 & 47.00 & 47.00 \\
Candidate disagreement & 57.00 & 58.50 & 57.75 & 57.75 \\
Residual-triggered & 61.50 & 62.25 & 62.25 & 62.00 \\
Expanded operator library & 76.50 & 77.25 & 77.25 & 77.00 \\
LLM-ACES & 72.75 & 73.50 & 74.25 & 73.50 \\
LLM-AutoSciLab & 75.50 & 76.25 & 77.00 & 76.25 \\
Falsification + independent & 78.75 & 79.50 & 80.25 & 79.50 \\
Joint edit--experiment & 89.00 & 90.25 & 89.25 & 89.50 \\
\bottomrule
\end{tabular}}
\end{table}

Table~\ref{tab:seed_source} resolves that stability across all eight methods. Every row reproduces the canonical $B=32$ mean, but the more informative comparison is between within-method variation and between-method separation: the joint policy's 1.25-point peak-to-peak range is small relative to its 10-point mean advantage over \indep{}. LLM-ACES spans 1.5 points and the independent falsification baseline also spans 1.5 points, so the magnitude of seed variation is comparable across learned methods while their mean recovery levels remain clearly separated.

This relationship between spread and effect size matters more than a single variance summary. Independent initialization changes the exact policy parameters, yet it does not erase the advantage associated with coupling structural and experimental decisions. The seed table therefore complements the ablation table: ablations show which algorithmic components create the gain, while seed replication shows that those components repeatedly produce the same ordering under separate training trajectories.

\begin{figure}[H]
\centering
\includegraphics[width=.90\linewidth]{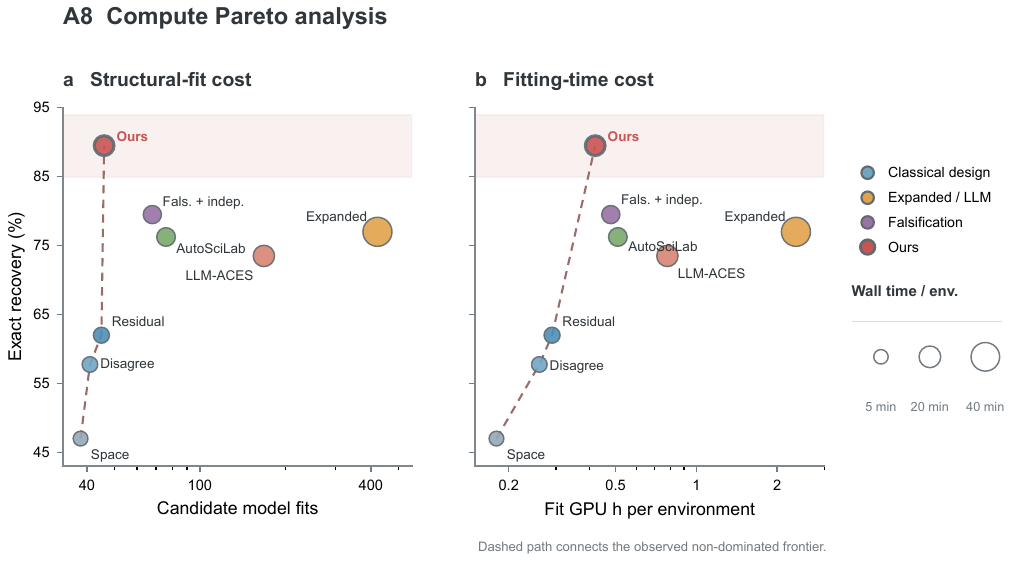}
\caption{\textbf{Discovery-time compute Pareto.} Recovery is plotted against candidate model fits and fitting GPU hours per environment. The joint policy lies on the observed non-dominated frontier.}
\label{fig:app_compute}
\end{figure}

Figure~\ref{fig:app_compute} places structural recovery on the same axes as the computational work needed to discover it. The expanded library evaluates 422 candidate models and consumes 2.35 fitting GPU hours per environment yet reaches 77.0\% recovery, whereas the joint policy reaches 89.5\% with 46 fits and 0.42 GPU hours. The expanded search therefore spends roughly nine times as many candidate fits and more than five times as much fitting GPU time while recovering fewer target structures. The difference is a search-allocation effect: breadth in model space becomes useful only when the system knows which structural alternatives are worth testing experimentally.

The Pareto geometry clarifies the role of the learned policy. Space filling and disagreement are computationally light but recover substantially fewer structures; large-library and LLM-driven methods explore more structural hypotheses but pay higher fit cost; the joint policy moves the frontier outward by using experiment-conditioned value to decide which hypotheses deserve fitting and execution. This is the computational counterpart of $\widehat J$: an edit receives search resources when it creates a prediction that can be exposed by an available experiment. Statistical efficiency and compute efficiency arise from the same selectivity principle.

\begin{table}[H]
\caption{Discovery-time compute at $B=32$. Teacher-search and policy-training compute are separate from these test-time discovery costs.}
\label{tab:compute_source}
\centering
\scriptsize
\resizebox{\linewidth}{!}{%
\begin{tabular}{lrrrr}
\toprule
Method & Proposer calls & Candidate fits & Fit GPU h / env. & Wall min / env. \\
\midrule
Space-filling & 7.0 & 38 & 0.18 & 5.6 \\
Candidate disagreement & 7.0 & 41 & 0.26 & 7.1 \\
Residual-triggered & 8.0 & 45 & 0.29 & 7.8 \\
Expanded operator library & 0.0 & 422 & 2.35 & 41.7 \\
LLM-ACES & 15.0 & 168 & 0.78 & 19.2 \\
LLM-AutoSciLab & 13.0 & 76 & 0.51 & 13.0 \\
Falsification + independent & 12.0 & 68 & 0.48 & 12.1 \\
Joint edit--experiment & 10.6 & 46 & 0.42 & 10.5 \\
\bottomrule
\end{tabular}}
\end{table}

Table~\ref{tab:compute_source} identifies the operational source of that frontier. Relative to \indep{}, the joint policy reduces candidate fits from 68 to 46 (32.4\%), fitting GPU time from 0.48 to 0.42 hours (12.5\%), and wall time from 12.1 to 10.5 minutes (13.2\%) while increasing recovery by 10 points. Compared with the expanded library, it uses roughly one ninth as many fits and one sixth as much fitting GPU time. The proposer-call count also falls from 12.0 to 10.6 relative to \indep{}, so the gain is visible in proposal, fit, and wall-clock accounting.

These resource changes explain how the joint objective alters search behavior. Independent selection can spend fit budget on edits whose corresponding experiments provide weak class separation; expanded-library search spends even more budget covering possibilities before their evidential value is known. Joint ranking makes prospective discriminability part of structural triage. The system therefore evaluates fewer candidate models because the experiment space supplies information about which edits can become testable, and it executes fewer real experiments because the retained edits are already paired with regimes that expose them. The two resource axes contract together because they are optimized through the same edit-conditioned evidence score.

\begin{figure}[H]
\centering
\includegraphics[width=.90\linewidth]{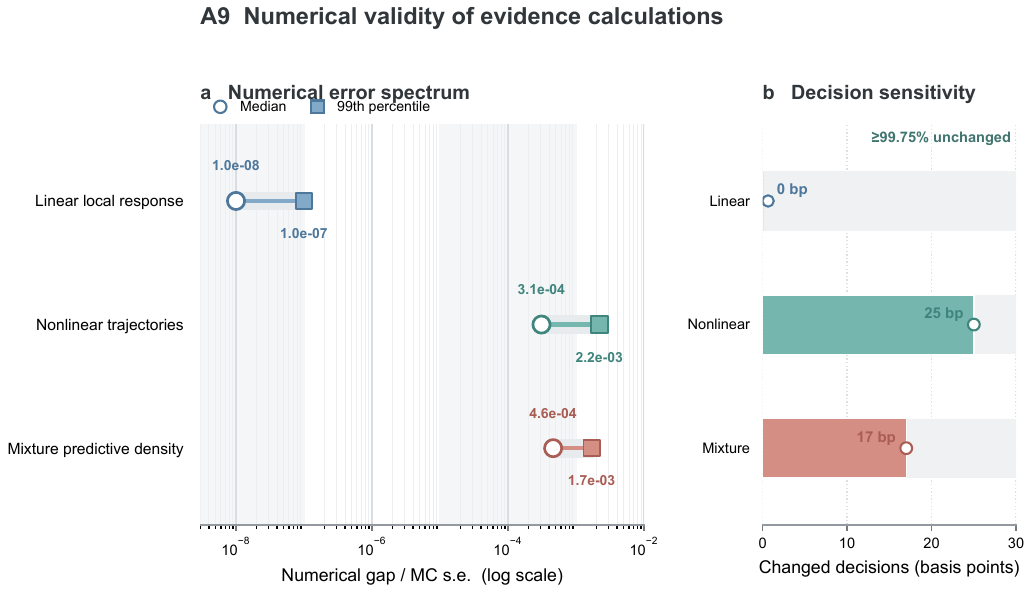}
\caption{\textbf{Numerical validity of evidence calculations.} Median / 99th-percentile numerical gaps are below $10^{-8}/10^{-7}$ for linear local responses, $3.1\times10^{-4}/2.2\times10^{-3}$ for nonlinear trajectories, and $4.6\times10^{-4}/1.7\times10^{-3}$ for mixture predictive densities. Tightening the numerical calculation changes at most 25 basis points of threshold decisions.}
\label{fig:app_numeric}
\end{figure}

Figure~\ref{fig:app_numeric} examines the numerical layer of the evidence gate, where continuous likelihood calculations ultimately determine a discrete revision action. Linear local-response calculations are effectively exact at the reported scale; nonlinear trajectories and mixture predictive densities show larger but still tightly concentrated errors, with 99th-percentile gaps of $2.2\times10^{-3}$ and $1.7\times10^{-3}$. Recomputing with tighter numerical settings changes at most 25 basis points of threshold decisions, leaving at least 99.75\% unchanged.

The important quantity is the distance between solver uncertainty and the decision boundary. The evidence trajectory can only support a stable revision semantics if numerical error is small enough that threshold crossings are determined by observations rather than by optimization tolerance or Monte Carlo variation. The decision-sensitivity panel directly evaluates this interface: tightening the computation changes only a tiny fraction of decisions even in the nonlinear and mixture settings. Numerical approximation therefore behaves as a small perturbation of the evidence process, allowing the empirical timing results in Figures~\ref{fig:case} and \ref{fig:app_traces} to be interpreted as properties of discovery trajectories rather than properties of solver precision.

\section{DiscoverPhysics, open-set stress, and additional traces}
\label{app:open}

\begin{figure}[H]
\centering
\includegraphics[width=.92\linewidth]{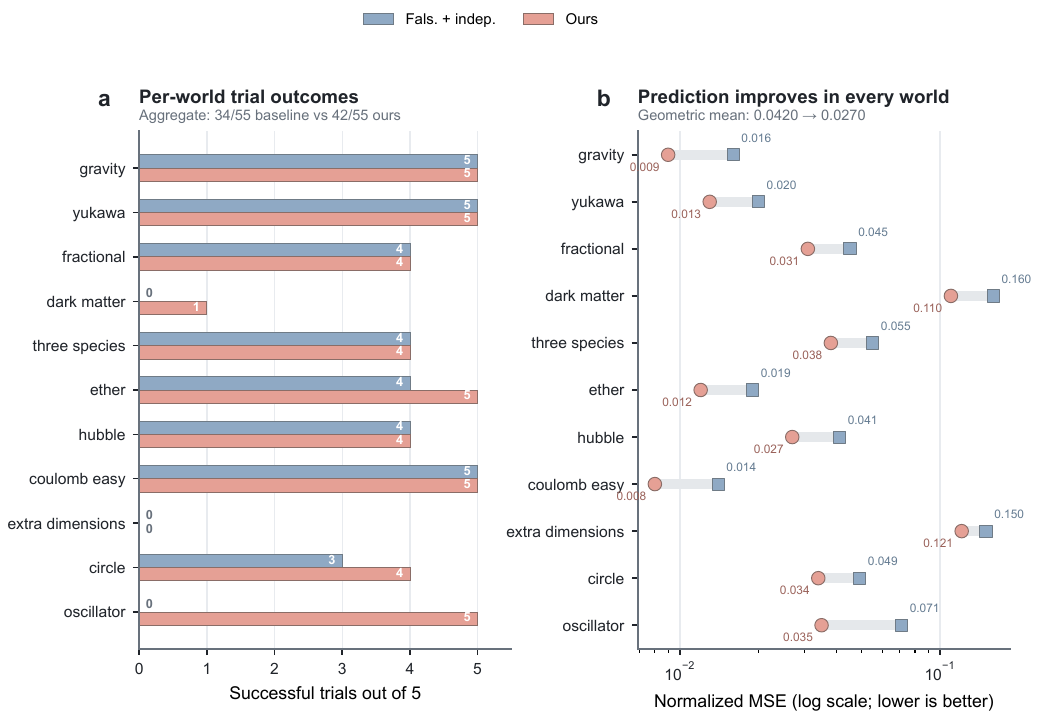}
\caption{\textbf{DiscoverPhysics per-world outcomes.} Across five matched trials per world, the baseline succeeds on 34/55 trials and \ours{} on 42/55. Prediction NMSE improves in every world shown; the geometric mean decreases from 0.0420 to 0.0270. Extra dimensions remains unsolved by both methods.}
\label{fig:app_dp}
\end{figure}

Figure~\ref{fig:app_dp} resolves the DiscoverPhysics result across 11 heterogeneous worlds. The joint method raises the aggregate pass count from 34/55 to 42/55 and improves normalized MSE in every world, so the gain appears both as better continuous prediction and as additional trials crossing the task-level success criterion. The eight additional successes are concentrated in dark matter, ether, circle, and especially oscillator, where the result changes from 0/5 to 5/5. Binary success therefore localizes the largest endpoint shifts to a small subset of worlds, while continuous prediction reveals a broader improvement across the full benchmark.

That difference between endpoint and trajectory is useful for interpreting transfer. In seven worlds the number of passes is unchanged, yet the recovered model predicts more accurately; the joint policy is improving the quality of the scientific model even when the improvement does not cross the benchmark's discrete success boundary. Oscillator contributes five of the eight additional passes, but it does not account for the universal NMSE direction. The result therefore has two layers: targeted edit--experiment selection can qualitatively change solvability in some worlds and quantitatively improve the recovered model in others. This is the external-world analogue of the controlled revision trajectory, where evidence first reshapes model quality and only then produces a discrete structural decision.

\begin{table}[H]
\caption{DiscoverPhysics pass counts and normalized MSE.}
\label{tab:dp_source}
\centering
\scriptsize
\resizebox{\linewidth}{!}{%
\begin{tabular}{lrrrr}
\toprule
World & Baseline pass/5 & Ours pass/5 & Baseline NMSE & Ours NMSE \\
\midrule
gravity & 5 & 5 & 0.016 & 0.009 \\
yukawa & 5 & 5 & 0.020 & 0.013 \\
fractional & 4 & 4 & 0.045 & 0.031 \\
dark matter & 0 & 1 & 0.160 & 0.110 \\
three species & 4 & 4 & 0.055 & 0.038 \\
ether & 4 & 5 & 0.019 & 0.012 \\
hubble & 4 & 4 & 0.041 & 0.027 \\
coulomb easy & 5 & 5 & 0.014 & 0.008 \\
extra dimensions & 0 & 0 & 0.150 & 0.121 \\
circle & 3 & 4 & 0.049 & 0.034 \\
oscillator & 0 & 5 & 0.071 & 0.035 \\
\midrule
Aggregate & 34/55 & 42/55 & 0.0420 & 0.0270 \\
\bottomrule
\end{tabular}}
\end{table}

Table~\ref{tab:dp_source} quantifies that two-level effect. The geometric-mean NMSE falls from 0.0420 to 0.0270, a 35.7\% relative reduction, and every one of the 11 rows moves in the same direction. The largest absolute errors remain in dark matter and extra dimensions, yet both also improve continuously, while lower-error worlds such as gravity, Yukawa, Coulomb, and Hubble move further toward the target dynamics. The improvement is therefore distributed across both difficult and already-competitive worlds.

The pass-count columns then reveal where these continuous gains become large enough to alter the discrete endpoint. Dark matter, ether, circle, and oscillator contribute new successes, while the other seven worlds retain the same pass count. This separation strengthens the interpretation of the benchmark: a pass is a thresholded summary of recovered model quality, whereas NMSE records the underlying predictive movement. Reading both together shows that the joint policy transfers as a model-improvement mechanism across all worlds and as a success-conversion mechanism where the improvement crosses the task criterion.

\begin{figure}[H]
\centering
\includegraphics[width=.92\linewidth]{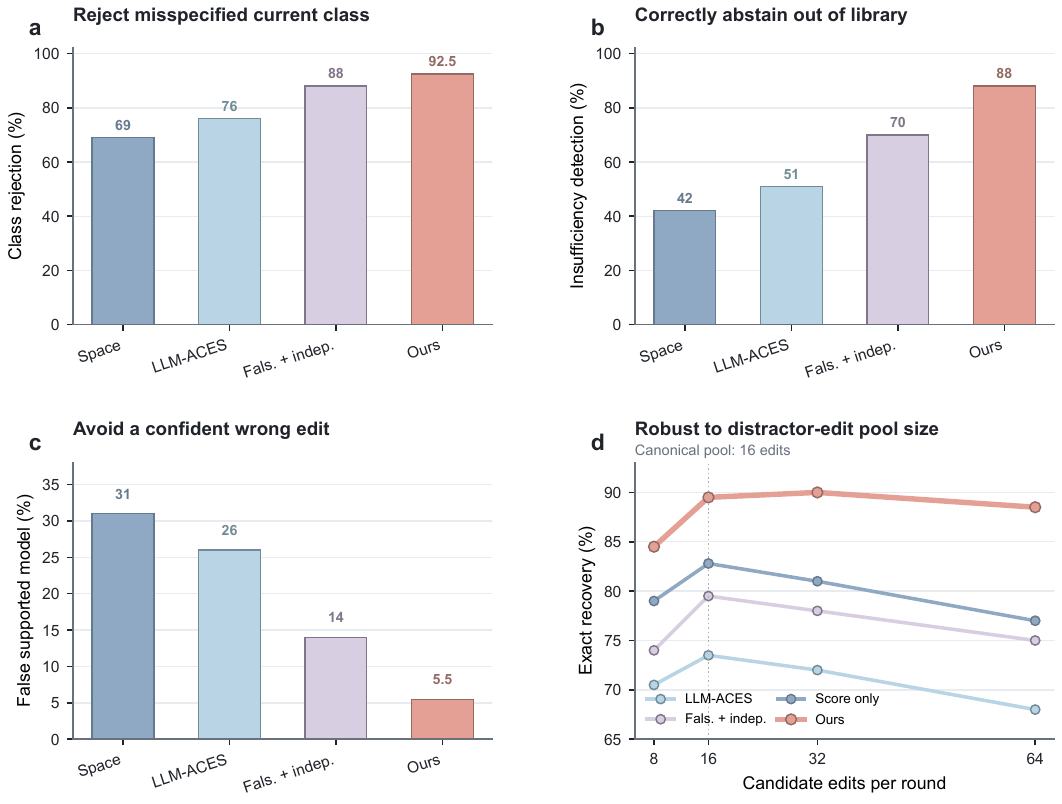}
\caption{\textbf{Open-set and distractor-edit stress.} (a) Class rejection on misspecified environments. (b) Correct abstention when the mechanism is outside the grammar. (c) False supported-model rate. (d) Exact recovery as the candidate-edit pool grows around the canonical 16-edit condition.}
\label{fig:app_open}
\end{figure}

Figure~\ref{fig:app_open} follows the discovery loop beyond ordinary in-grammar recovery. On mechanisms excluded from the edit library, the joint policy rejects the current class in 92.5\% of cases, identifies library insufficiency in 88.0\%, and supports an incorrect in-library model in only 5.5\%. The 4.5-point gap between class rejection and insufficiency detection is far smaller than for \indep{} (18 points). Evidence against the current model class is therefore translated into the correct structural endpoint much more consistently when edit selection and experiment selection remain coupled.

Panel (d) tests the complementary pressure: the policy is given increasingly many in-grammar alternatives. Recovery rises from 84.5\% with eight edits to 90.0\% with 32, then remains 88.5\% with 64. A larger neighborhood initially helps by exposing useful structural moves, while a very large neighborhood adds enough distractors to lower recovery slightly. Taken together, the open-set and distractor panels describe the same decision geometry from opposite sides. The policy must exploit genuine structural breadth when a supported edit exists and preserve an insufficiency outcome when no available edit explains the evidence. The learned pair score supplies the routing signal between these two regimes.''

\begin{table}[H]
\caption{Out-of-library comparison on 200 mechanisms intentionally excluded from the edit grammar.}
\label{tab:open_source}
\centering
\small
\begin{tabular}{lrrrr}
\toprule
Method & Class reject & Detect insufficiency & False support & Exp. used \\
\midrule
Space + common diagnostic & 69.0 & 42.0 & 31.0 & 29.8 \\
LLM-ACES + common diagnostic & 76.0 & 51.0 & 26.0 & 26.4 \\
Falsification + independent & 88.0 & 70.0 & 14.0 & 23.5 \\
Joint edit--experiment & \textbf{92.5} & \textbf{88.0} & \textbf{5.5} & \textbf{21.9} \\
\bottomrule
\end{tabular}
\end{table}

Table~\ref{tab:open_source} decomposes the open-set decision into its three logical stages: rejecting the current class, recognizing that the retained edit library does not explain the discrepancy, and avoiding a false supported model. Relative to \indep{}, joint selection raises insufficiency detection by 18 points, reduces false support by 8.5 points, and reaches the terminal decision with 1.6 fewer experiments on average. The movement of all three quantities in the same direction shows that earlier stopping coincides with a more accurate structural conclusion and a lower experimental budget.

The difference between class rejection and insufficiency detection is especially revealing. A rejection event says that the current class is inadequate; it does not specify which neighboring class should replace it. The small 4.5-point gap for the joint policy indicates that the post-rejection structural search usually preserves the meaning of the evidence and reaches the intended open-set endpoint. The 18-point gap for independent selection shows how much information can be lost between detecting misspecification and choosing the structural response. This table therefore isolates a core function of joint pairing: it connects falsification evidence to the next representational action.

\begin{table}[H]
\caption{Candidate-edit distractor stress. Canonical controlled evaluation uses 16 candidate edits per round.}
\label{tab:distractor_source}
\centering
\small
\begin{tabular}{rrrrr}
\toprule
Candidate edits & LLM-ACES & Fals.+indep. & Score only & Ours \\
\midrule
8 & 70.5 & 74.0 & 79.0 & 84.5 \\
16 & 73.5 & 79.5 & 82.8 & 89.5 \\
32 & 72.0 & 78.0 & 81.0 & 90.0 \\
64 & 68.0 & 75.0 & 77.0 & 88.5 \\
\bottomrule
\end{tabular}
\end{table}

Table~\ref{tab:distractor_source} reveals how learned selectivity scales with the size of the structural neighborhood. The joint policy leads at every pool size, and its margin over \indep{} grows from 10.5 points with eight candidates to 13.5 points with 64 candidates. Recovery peaks at 90.0\% with 32 edits and declines slightly to 88.5\% at 64, while every competing method also declines at the largest pool. The pool-size sweep therefore has a non-monotone structure: additional breadth is useful until the density of structurally plausible but experimentally weak alternatives becomes large enough to dilute selection.

The widening margin at 32 and 64 candidates identifies the role of the experiment dimension. When the pool is small, structural scores alone already remove many poor choices. As the pool grows, several edits can look plausible under the existing data, and their relative value increasingly depends on whether an available intervention can separate them from the current class. Joint ranking adds exactly that second coordinate. The policy therefore gains comparative value as structural ambiguity grows, because the experiment space becomes a filter for deciding which of many syntactically plausible edits are empirically resolvable.

\begin{figure}[H]
\centering
\includegraphics[width=.94\linewidth]{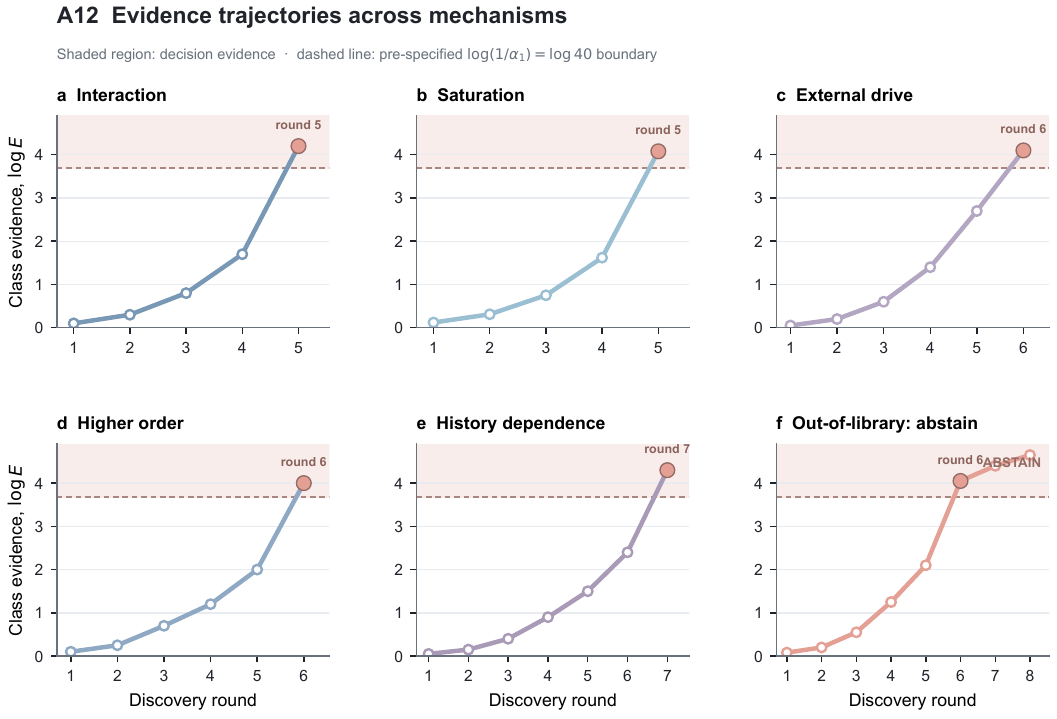}
\caption{\textbf{Evidence trajectories across mechanisms.} The interaction and saturation cases cross the $\log 40$ boundary at round 5, external-drive and higher-order cases at round 6, history dependence at round 7, and the out-of-library trajectory rejects the current class before terminating with an insufficiency decision.}
\label{fig:app_traces}
\end{figure}

Figure~\ref{fig:app_traces} shows that the revision mechanism in Figure~\ref{fig:case} recurs as a shared trajectory shape across mechanism families. Interaction and saturation cross the evidence boundary at round 5, external drive and higher-order dynamics at round 6, and history dependence at round 7. These different crossing times are a dynamic signature of mechanism-specific distinguishability: edits whose consequences are exposed directly in the response geometry generate decisive evidence earlier, while temporally mediated mechanisms require a longer sequence of targeted probes.

The out-of-library trajectory is the key structural contrast. Its evidence path still crosses the same class-rejection boundary, establishing that the current class cannot explain the observations, but the next action is abstention because no retained edit receives sufficient support. The figure therefore separates two decisions that are easy to conflate in a discovery system: whether the current theory survives and what representational move should follow if it does not. Across all six traces, the common sequence is experiment selection $\rightarrow$ class-level evidence $\rightarrow$ structural routing. The endpoint changes with the supported neighborhood, while the evidence semantics remain shared. This is the temporal realization of the architecture in Figure~\ref{fig:overview}.

\section{Additional numeric source tables}
\label{app:numeric}

\begin{table}[H]
\caption{Budget curve source values: exact recovery (\%).}
\centering
\scriptsize
\resizebox{\linewidth}{!}{%
\begin{tabular}{rrrrrrrrr}
\toprule
$B$ & Space & Disagree & Residual & Expanded & LLM-ACES & AutoSciLab & Fals.+indep. & Ours \\
\midrule
8 & 22.0 & 27.5 & 30.0 & 37.0 & 38.0 & 40.5 & 43.0 & 56.0 \\
16 & 34.5 & 42.0 & 46.0 & 58.5 & 58.0 & 61.0 & 64.0 & 76.5 \\
32 & 47.0 & 57.75 & 62.0 & 77.0 & 73.5 & 76.25 & 79.5 & 89.5 \\
64 & 58.0 & 68.0 & 72.5 & 86.0 & 82.5 & 84.5 & 87.5 & 94.5 \\
\bottomrule
\end{tabular}}
\label{tab:budgetsrc}
\end{table}

Table~\ref{tab:budgetsrc} exposes the sample-efficiency signature of joint selection across the full experiment-budget curve. The advantage over \indep{} is largest when evidence is scarce (+13.0 points at $B=8$ and +12.5 at $B=16$), then narrows to +10.0 and +7.0 points as the budget grows. This contraction is expected when additional experiments allow less selective strategies to revisit regimes they failed to prioritize early, yet the joint policy remains ahead at every budget.

The cross-budget comparison is stronger than the within-budget margins alone. At $B=32$, the joint policy reaches 89.5\% recovery, exceeding the 87.5\% achieved by \indep{} at $B=64$. The same structural endpoint is therefore reached with substantially fewer physical interventions. This is the empirical meaning of ``more discovery per experiment'': joint pairing changes the order in which informative regimes are visited, so the policy accumulates decisive class evidence before a method that chooses edits and experiments independently can compensate through additional trials. The budget curve turns the main architectural choice into a direct sample-efficiency statement.

\begin{table}[H]
\caption{Transfer source values: exact recovery (\%). The held-out primitive remains expressible by the edit grammar.}
\label{tab:transfer_source}
\centering
\small
\begin{tabular}{lrrr}
\toprule
Condition & LLM-ACES & Fals.+indep. & Ours \\
\midrule
Seen composition & 78.0 & 82.5 & 90.5 \\
Unseen primitive combination & 66.5 & 72.0 & 84.0 \\
Held-out primitive & 58.0 & 68.0 & 79.2 \\
Parameter extrapolation & 62.0 & 68.0 & 80.5 \\
Shifted experiment cost & 61.0 & 69.5 & 82.0 \\
\bottomrule
\end{tabular}
\end{table}

Table~\ref{tab:transfer_source} tests what the policy has learned about discovery. On seen compositions, the joint method exceeds \indep{} by 8.0 points. Under unseen primitive combinations the margin grows to 12.0 points; under a held-out but still expressible primitive it is 11.2 points; parameter extrapolation and shifted experiment cost each produce a 12.5-point margin. The ordering is therefore preserved as both mechanism composition and the economics of experimentation change.

The edit grammar remains executable in these tests, while the required edit--experiment relation changes under new combinations, parameter ranges, or costs. The larger margins under several shifted conditions are consistent with a relational policy: it has learned that the value of an experiment depends on the structural alternative being tested, and that relation can be recomputed when the environment or cost landscape changes. Transfer therefore occurs at the level of edit--experiment compatibility rather than at the level of memorized experimental actions.

\begin{table}[H]
\caption{Paired binary outcomes for external ODE revision.}
\label{tab:ode_pairs}
\centering
\small
\begin{tabular}{lrrrrr}
\toprule
Dataset & Both success & Ours only & Baseline only & Both fail & McNemar $p$ \\
\midrule
ODEBench & 35 & 9 & 1 & 18 & 0.0215 \\
ODEBase & 35 & 9 & 1 & 15 & 0.0215 \\
\bottomrule
\end{tabular}
\end{table}

Table~\ref{tab:ode_pairs} resolves the public-ODE percentages at the paired system level. Both ODEBench and ODEBase contain nine systems solved only by the joint policy and one solved only by \indep{}, while the number of systems failed by both differs across the two collections. This repeated 9:1 discordant-pair asymmetry yields the same exact McNemar $p=0.0215$ despite different dataset sizes.

The paired view changes the interpretation of the external result. A marginal recovery increase could arise from different subsets of easy and difficult systems in each method, but the discordant cells identify the direction of change on the same systems: nine systems move from failure under independent selection to success under joint pairing for every one moving in the opposite direction. The repetition of that asymmetry in both libraries links the aggregate improvement to per-system revision behavior. Coupled selection is therefore associated with a systematic expansion of the set of external systems for which the discovery loop reaches a supported structural revision.

\end{document}